\documentclass{article}

\usepackage[preprint]{neurips_2024}

\usepackage[dvipsnames]{xcolor}

\usepackage{relsize}

\usepackage{multirow}
\usepackage{xcolor, colortbl}

\definecolor{darkred}{rgb}{0.7,0.1,0.1}
\definecolor{darkgreen}{rgb}{0.1,0.6,0.1}
\definecolor{cyan}{rgb}{0.7,0.0,0.7}
\definecolor{otherblue}{rgb}{0.1,0.4,0.8}
\definecolor{maroon}{rgb}{0.76,.13,.28}
\definecolor{burntorange}{rgb}{0.81,.33,0}
\definecolor{apricot}{rgb}{0.98, 0.81, 0.69}

\usepackage{pifont}
\newcommand{\cmark}{\color{darkgreen}{\ding{51}}}
\newcommand{\xmark}{\color{darkred}{\ding{55}}}

\usepackage{caption}
\usepackage{booktabs}
\usepackage{arydshln}

\newcommand{\E}{\mathbb{E}}

\newcommand\blfootnote[1]{%
  \begingroup
  \renewcommand\thefootnote{}\footnote{#1}%
  \addtocounter{footnote}{-1}%
  \endgroup
}

\newcommand{\eclipse}{\textit{ECLIPSE}}
\newcommand{\ours}{\texttt{$\lambda$\textit{-ECLIPSE}}}

\newcommand{\nresults}[1]{\textsubscript{{\color{BrickRed}\textbf{#1}}}}

\usepackage{subcaption}
\usepackage[utf8]{inputenc} 
\usepackage[T1]{fontenc}    
\usepackage{url}            
\usepackage{booktabs}       
\usepackage{amsfonts}       
\usepackage{nicefrac}       
\usepackage{microtype}      
\usepackage{xcolor}         
\usepackage{graphicx}
\usepackage{amsmath,amsfonts}

\usepackage[pagebackref,breaklinks,colorlinks]{hyperref}
\hypersetup{linkcolor=red,anchorcolor=blue,citecolor=cyan,filecolor=blue,urlcolor=blue}

\title{\ours: Multi-Concept Personalized Text-to-Image Diffusion Models by Leveraging CLIP Latent Space} 

\author{%
  Maitreya Patel\textsuperscript{*}$^\dagger$ \quad \quad Sangmin Jung\textsuperscript{*} \quad \quad Chitta Baral \quad \quad  Yezhou Yang \\
  Arizona State University\\
  \url{https://eclipse-t2i.github.io/Lambda-ECLIPSE/}
}

\begin{document}
\maketitle

\blfootnote{* indicates equal contribution, $\dagger$ corresponding author: \texttt{maitreya.patel@asu.edu}}

\begin{center}
\vspace{-7ex}
    \centering
    \captionsetup{type=figure}
    \includegraphics[width=\textwidth]{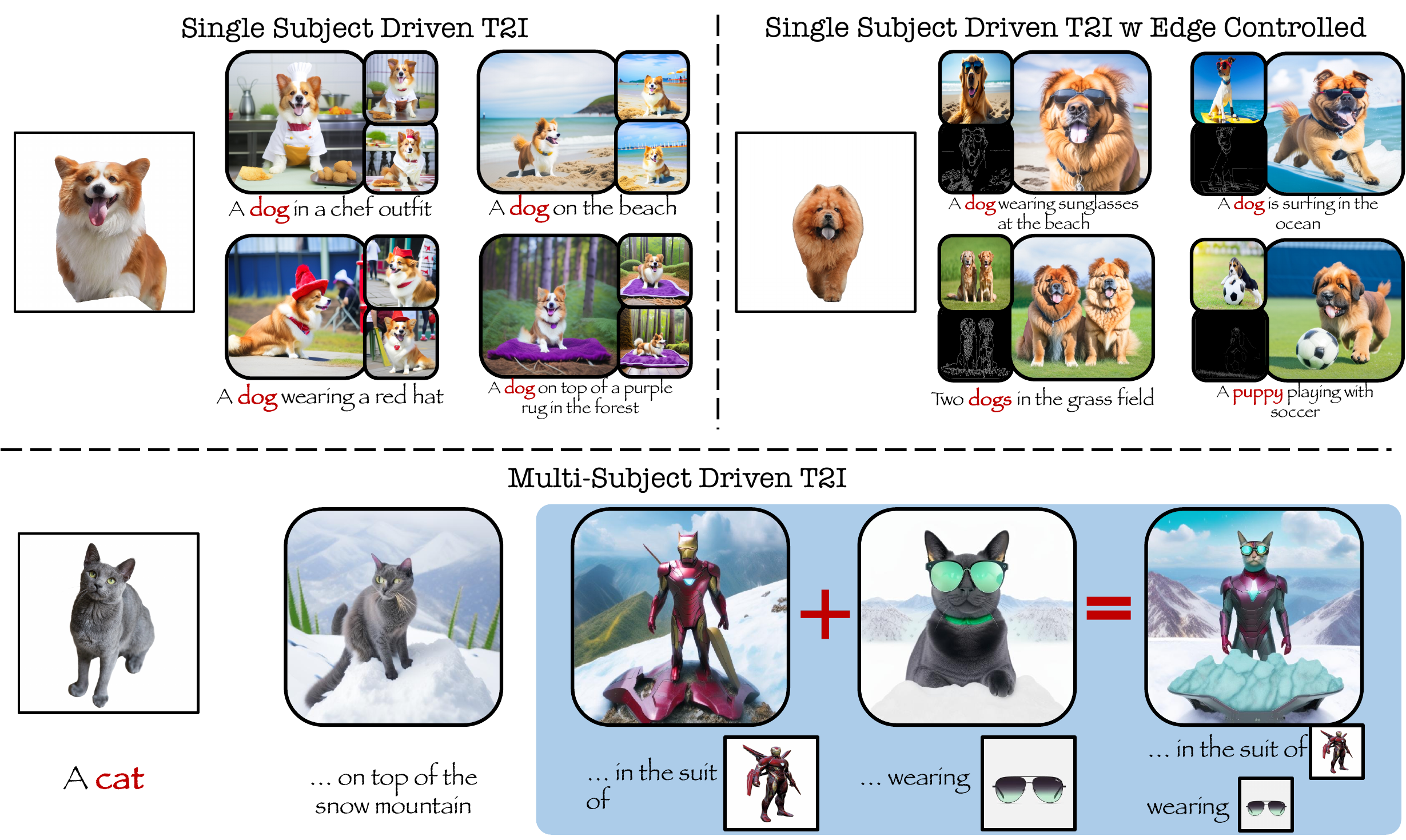}
    \captionof{figure}{
    \ours~ can estimate subject-specific image embeddings while maintaining the balance between concept and composition alignment in a resource-efficient way. 
    }
    \label{fig:single_subject_quantitative}
\end{center}

\begin{abstract}

Despite the recent advances in personalized text-to-image (P-T2I) generative models, it remains challenging to perform finetuning-free multi-subject-driven T2I in a resource-efficient manner.
Predominantly, contemporary approaches, involving the training of Hypernetworks and Multimodal Large Language Models (MLLMs), require heavy computing resources that range from 600 to 12300 GPU hours of training.
These subject-driven T2I methods hinge on Latent Diffusion Models (LDMs), which facilitate T2I mapping through cross-attention layers. 
While LDMs offer distinct advantages, P-T2I methods' reliance on the latent space of these diffusion models significantly escalates resource demands, leading to inconsistent results and necessitating numerous iterations for a single desired image.
In this paper, we present \ours, an alternative prior-training strategy that works in the latent space of a pre-trained CLIP model without relying on the diffusion UNet models.
\ours~leverages the image-text interleaved pre-training for fast and effective multi-subject-driven P-T2I.
Through extensive experiments, we establish that \ours~surpasses existing baselines in composition alignment while preserving concept alignment performance, even with significantly lower resource utilization.
\ours~performs multi-subject driven P-T2I with just 34M parameters and is trained on a mere 74 GPU hours.
Additionally, \ours~demonstrates the unique ability to perform multi-concept interpolations.
\end{abstract}

\section{Introduction}
The field of text-to-image (T2I) diffusion models has recently witnessed remarkable advancements, achieving greater photorealism and enhanced adherence to textual prompts.
This has catalyzed the emergence of diverse applications, notably subject-driven personalized T2I (P-T2I) models. 
In particular, this encompasses the intricate task of learning and reproducing novel visual concepts or subjects in varied contexts requiring high concept and compositional alignment.
The complexity escalates further when multi-subject personalization is desired.

Early works employed concept-specific optimization strategies involving fine-tuning certain parameters within T2I diffusion models~\cite{gal2022image, ruiz2023dreambooth, kumari2023multi, tumanyan2023plug, gal2023encoder}. 
Although these methods achieve state-of-the-art (SOTA) performance, they struggle with generalization and are time-intensive. 
Contemporary research is pivoting towards fast personalization techniques.
Within this paradigm, there are two types of popular approaches:
1) Methods that involve training hypernetworks and integrating new layers or parameters within pre-trained diffusion UNet models~\cite{Wei2023ELITEEV,ye2023ip,tewel2023key,shah2023ziplora, ruiz2023hyperdreambooth}, and 
2) MLLM-based learning of prior models that focuses on leveraging text-latent space of frozen diffusion UNet model~\cite{pan2023kosmos, sun2023generative}.

The hypernetwork-based strategy achieves single-concept customization but has not been extended to multi-concepts. 
Moreover, when combined with additional control (i.e. Canny edge map), they struggle to maintain the concept alignment ($\sim$30\% drop in performance; Section~\ref{sec:experiments}) and strongly favor the edge map.
At the same time, MLLM-based approaches can perform fast multi-concept customization but require heavy computing resources.
In Table~\ref{tab:previous_methods_overview}, we provide the overview of various single and multi-concept customization methodologies in terms of total parameters, iterations, and GPU hours required to train the models.
It can be observed that multi-concept customization methodologies further increase the resource requirements.
For instance, Kosmos-G~\cite{pan2023kosmos} consumes 18x more resources than IP-Adapter~\cite{ye2023ip}.
And Emu2~\cite{sun2023generative} requires training of 19x more parameters compared to Kosmos-G.
Hence, despite MLLMs' seemingly useful scenarios it is not viable to blindly train them.
Therefore, in this work, we focus on answering one question: \texttt{What are the alternative methodology and design choices we can make to improve resource efficiency?}

Upon further investigation, we find that most subject-driven T2I approaches build upon variants of the Latent Diffusion Model (LDM)~\cite{rombach2022high}, specifically Stable Diffusion models. 
These LDMs employ cross-attention layers to condition diffusion models with text embeddings, necessitating a mapping of target subject images to latent spaces compatible with the diffusion models at the prior training stage.
This is also known as score distillation instruction tuning for MLLMs~\cite{pan2023kosmos}.
As there is no choice but to learn this text-to-image diffusion latent space, it involves backpropagation through the entire diffusion model often comprising over a billion parameters, contributing to the inefficiency of existing P-T2I methods. 

To improve the resource efficiency for multi-concept image generation, we present \ours\footnote{The designation \ours~is inspired by its conceptual alignment with the $\lambda$-calculus. In this context, the \ours~model functions similarly to a functional abstraction within $\lambda$-calculus, where it effectively binds variables. These variables, in our case, represent novel visual concepts that are integrated through composition prompts. Here, \eclipse~indicates our architecture design choice.}, which leverages the properties of UnCLIP T2I models (e.g. DALL-E 2~\cite{ramesh2022hierarchical} and Kandinsky v2.2~\cite{razzhigaev2023kandinsky}) and performs P-T2I in the compressed latent space.
Specifically, unlike previous MLLM-based methodologies, \ours~aligns the output space of priors with CLIP vision space instead of the CLIP text space.
\ours~takes multiple images and text instructions as input and estimates the respective vision embeddings, which can be used by the frozen diffusion UNet model from the UnCLIP stack to generate the resulting image.
This elevates the training time dependencies on diffusion models for P-T2I; significantly contributing to the resource efficiency.
Additionally, as diffusion or MLLM-based priors are still compute heavy due to a huge number of parameters and slower convergence, we build upon \eclipse~\cite{patel2023eclipse} and SEED~\cite{ge2023making}, which shows that text-to-image mapping can be optimized through contrastive pre-training.
Here, we select \eclipse~as preferred choice of prior architecture for best efficiency.
At last, we propose a subject-driven instruction tuning task based on the image-text interleaved data as a pre-training strategy. 
This involves creating 2 million high-quality image-text pairs, where text embeddings linked to subjects are substituted with the respective image embeddings, which in return are considered as input to the \ours.
While \ours~can be plugged with these pre-trained methods, we explore the possibility of \ours~to incorporate Canny edge as an additional control to synergetically work with subject-driven image generation tasks. 
Figure~\ref{fig:single_subject_quantitative} provides the overview of \ours~capabilities.

Overall, we propose \ours~as an initial attempt to motivate future works on designing resource-efficient solutions for MLLM-based approaches. 
We summarize our main contributions as follows:
1) We introduce a training-time diffusion-independent, UnCLIP-based prior learning strategy for enabling efficient and fast multi-subject customization.
2) Extensive experiments on Dreambench, Multibench, and ConceptBed reveal that \ours~(34M parameter model) trained on a mere 74 GPU hours can achieve competitive performance to that of big counterparts (having 2B-37B parameters) and improve text-composition alignment.  
3) At last, \ours~inherits the smooth CLIP latent space. This allows us to perform the seamless transition between multi-concept generated images. 
\section{Related Works}
\label{sec:related_works}

\begin{table}[!t]
    \centering
    \caption{\textbf{A quick overview of previous works on P-T2I.} Our method is the first to offer multi-concept-driven generation without depending on diffusion UNet models (except for inference). We provide the extended overview table in the appendix.}
    
    \resizebox{1\columnwidth}{!}{
        \begin{tabular}{>{\columncolor[gray]{0.95}}lccccccc}
            \toprule
            
        \textbf{Method} & \textbf{Multi} & \textbf{Finetuning} & \textbf{Diffusion} & \textbf{Total opt.} & \textbf{Training} & \textbf{Dataset} & \textbf{GPU} \\
         & \textbf{Concepts} & \textbf{Free} & \textbf{Free} & \textbf{params} & \textbf{Steps} & \textbf{Size}     &\textbf{Hours} \\

            \midrule

            Textual Inversion~\cite{gal2022image}  & \xmark & \xmark & \xmark & 768 & 5000 & - & 1 \\

            DreamBooth~\cite{ruiz2023dreambooth}  & \xmark & \xmark & \xmark & 0.9B & 800 & - & 0.2 \\

            ELITE~\cite{Wei2023ELITEEV} & \xmark & \cmark & \xmark & 77M & 135K & 125K & 336 \\

            BLIP-Diffusion~\cite{li2023blip} & \xmark & \cmark & \xmark & 1.5B & 500K & 129M & 2304 \\

            IP-Adapter~\cite{ye2023ip} & \xmark & \cmark & \xmark & 22M & 1M & - & 672 \\

            \hdashline
            
            Custom Diffusion~\cite{kumari2023multi}  & \cmark & \xmark & \xmark & 57M & 500 & - & 0.1 \\

            Subject-Diffusion~\cite{ma2023subject} & \cmark & \cmark & \xmark & 252M & 300K & 76M & -\\

            Kosmos-G~\cite{pan2023kosmos} & \cmark & \cmark & \xmark & 1.9B & 800K & 200M & 12300 \\

            Emu-2~\cite{sun2023generative} & \cmark & \cmark & \xmark & 37B & 70K & 162M & - \\

            \midrule 
            
            \rowcolor{apricot!50}\textbf{\ours~(ours)} & \cmark & \cmark  & \cmark & 34M & 100K & 2M & 74 \\
            
            \bottomrule
        \end{tabular}
    }
    \label{tab:previous_methods_overview}
\end{table}

\paragraph{\textbf{Text-to-Image Generative Models.}}
Pioneering efforts in image generation, notably DALL-E~\cite{ramesh2021zero} and CogView~\cite{ding2021cogview}, leveraged autoregressive models to achieve significant results. 
Recent advancements predominantly feature diffusion models, acclaimed for their high image fidelity and diversity in text-to-image (T2I) generation. 
A notable example is Stable Diffusion, which builds upon the Latent Diffusion Model (LDM)~\cite{rombach2022high} and excels in semantic and conceptual understanding by transitioning training to latent space.
Imagen~\cite{saharia2022photorealistic}, Pixart-$\alpha$~\cite{chen2023pixart}, and DALL-E 3~\cite{betker2023improving} propose using a large T5 language model to improve language understanding.
DALL-E 2~\cite{ramesh2022hierarchical} along with its UnCLIP variation models such as Kandinsky~\cite{razzhigaev2023kandinsky} and Karlo~\cite{kakaobrain2022karlo-v1-alpha}, uses a diffusion prior and diffusion UNet modules to generate images using the pre-trained CLIP~\cite{radford2021learning} model.

\paragraph{\textbf{Personalized T2I Methods.}}
Approaches like Textual Inversion~\cite{gal2022image}, DreamBooth~\cite{ruiz2023dreambooth}, and Custom Diffusion~\cite{kumari2023multi} focus on training specific parameters to encapsulate visual concepts.
LoRA~\cite{hu2021lora} and Perfusion~\cite{tewel2023key} target efficient fine-tuning adjustments, particularly rank 1 modifications.
However, these methods are constrained by their requirement for concept-specific tuning.
ELITE~\cite{Wei2023ELITEEV} was the first approach addressing fast customized generation for single-subject T2I.
BLIP-Diffusion~\cite{li2023blip} adapts the BLIP2 encoder~\cite{li2023blip2}, training approximately 1.5B parameters to enable zero-shot, subject-driven image generation.
IP-Adapter introduces a decoupled cross-attention mechanism, negating the need to train the foundational UNet model by permitting fine-tuning of a reduced number of 22M parameters.

Mix-of-Show~\cite{gu2023mix} and Zip-LoRA~\cite{shah2023ziplora} train individual concepts and then combine them to generate multiple subjects. 
Break-A-Scene~\cite{avrahami2023break} shows multi-concept capability but requires single images containing diverse objects.
Subject Diffusion~\cite{ma2023subject} creates a high-quality dataset and presents the precision control for fast personalized multi-subject image generation.
Kosmos-G and Emu2~\cite{sun2023generative}, akin to Subject-Diffusion~\cite{ma2023subject}, employs a Multimodal Large Language Model (MLLM) for text-image embedding alignment, though it necessitates extensive parameter optimization (1.9B-37B). 
These multi-subject P-T2I methods are not only demanding in terms of parameters but also depend on a massive number of frozen parameters of the diffusion UNet model, increasing training computational loads. 
In contrast, our model, \ours, forgoes test-time fine-tuning and training-time reliance on the diffusion UNet model for single and multi-concept, control-guided P-T2I, positioning it as a resource-efficient solution.

At last, methods like GLIGEN~\cite{li2023gligen}, ControlNet~\cite{zhang2023adding}, and UniControl~\cite{qin2023unicontrol} incorporate additional controls (i.e., edge map, depth, segmentations) into the diffusion model to generate the desired images.
BLIP-Diffusion, IP-Adapter, and Kosmos-G can leverage such pre-trained controls. 
However, in many scenarios, these controls are too strong, making generated images lose subject-specific details.
We show that \ours~learns to balance the edge map, subjects, and composition.
We offer a more comprehensive review of related works in the appendix.

\section{Method}

\begin{figure*}[!t]
    \includegraphics[width=\textwidth]{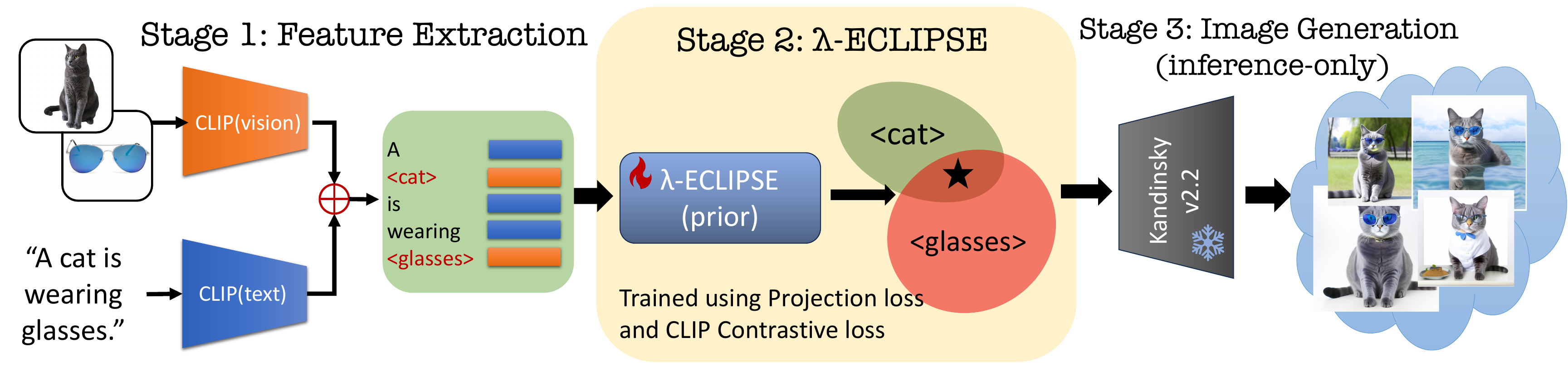}
    \caption{
    \textbf{This figure illustrates the three stages of the \ours~pipeline.}
    1) Create the image-text interleaved features using frozen CLIP. 
    2) Pre-train the \ours~(34M parameters) using Eq.~\ref{eq:eclipse}, which ensures the mapping to the desired latent space given the image-text interleaved data.
    3) During inference, the frozen Kandinsky v2.2 diffusion UNet model takes the output from the \ours~ and generates the image. 
    }
    \label{fig:pipeline}
\end{figure*}

In this section, we introduce \ours, our approach to multi-subject personalized text-to-image generation. 
Our method combines the contrastive text-to-image strategy from \eclipse~with the novel image-text interleaved pretraining strategy, notably omitting the need for explicit diffusion modeling. 
Our approach mainly capitalizes on the efficient utilization of the CLIP latent space. 
Figure~\ref{fig:pipeline} outlines the end-to-end framework.

The primary objective of \ours~is to facilitate single and multi-subject P-T2I generation processes, accommodating edge maps as conditional guidance.
Initially, we detail the problem formulation and elaborate on the UnCLIP stack design of the \ours.
Subsequently, we delve into the image-text interleaved training methodology. 
This fine-tuning process enables the \ours~to harness semantic correlations between CLIP image and text latent spaces while preserving subject-specific visual features. 

\subsection{Text-to-Image Prior Mapping}

\begin{figure}[!t]
    \includegraphics[width=\linewidth]{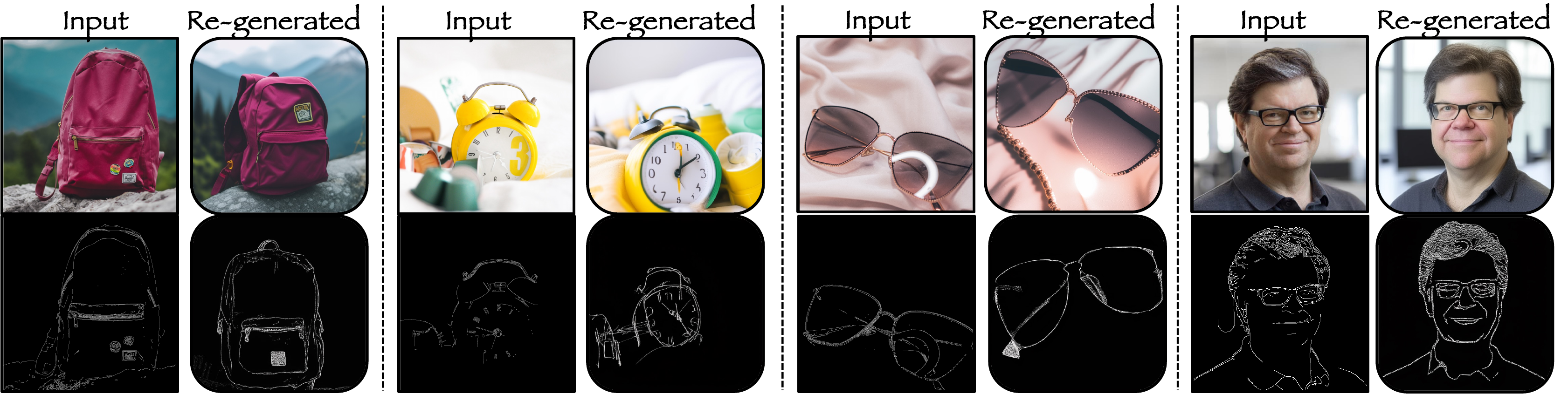}
    \caption{
    \textbf{CLIP(vision) features capture the semantics and fine-grained visual details.} Each input is given as input to the Kandinsky v2.2 and re-generated from the decoder. (Top: Real-images, Bottom: Canny edge)
    }
    \label{fig:motivation}
\end{figure}

In the UnCLIP T2I models, the objective of the text-to-image prior model ($f_\theta$) is to establish a proficient text-to-image embedding mapping. 
This model is designed to adeptly map textual representations to their corresponding visual embeddings, denoted as ($f_\theta : z_y \rightarrow z_x$), where $z_{x/y}$ represent the embeddings for images and text, respectively. 
The visual embedding predictions ($\hat{z_x} = f_\theta(z_y)$) are then effectively utilized by the diffusion image generators ($h_\phi$), which are inherently conditioned on these vision embeddings ($h_\phi : z_x \rightarrow x$).
In our experiments, we utilize the Kandinsky v2.2 diffusion UNet model as $h_\phi$.

As shown in Figure~\ref{fig:motivation}, the CLIP vision encoder preserves the necessary features in $z_x$ required to reconstruct the input image and achieves a similar concept alignment score (\texttt{DINO: 0.66}) as finetuning-based DreamBooth method~\cite{ruiz2023dreambooth}. 

Our goal is to accurately estimate the image embedding $\hat{z}_x$, incorporating the subject representations, thereby eliminating reliance on $h_\phi$ during training.
Existing LDM-based P-T2I methods are limited by the LDM's singular module approach ($h_\phi: z_y \rightarrow x$).
Consequently, mastering the latent space of $h_\phi$ becomes essential for effective P-T2I for the baseline methodologies, which limits the previous methodologies.

We propose a new mapping function, $f_{\theta}$, which processes text representations ($z_y$) alongside subject ($x_k$) specific visual representations ($z_{x_k}$), to derive an image embedding that encapsulates both text prompts and subject visuals ($\hat{z}_x$). 
The challenge lies in harmonizing $z_{x_k}$ and $z_y$ within $f_{\theta}: (z_y, z_{x_k}) \rightarrow \hat{z}_x$, ensuring alignment while preventing overemphasis on either aspect, as this could compromise composition alignment.
To address this, we employ the contrastive pre-training strategy after \cite{patel2023eclipse}:

\begin{equation}
    \begin{split}
        \label{eq:eclipse}
            \mathcal{L}_{prior} = 
            \underset{
                \substack{
                    \epsilon \sim \mathcal{N}(0,I) \\
                    z_y, z_{x_k}
                }
            }{\E} \Big[ || z_x - f_\theta(\epsilon, z_y, z_{x_k})||_2^2 \Big]
             - \frac{\lambda}{N} \sum_{i=0}^{N} \log \frac{\exp(\langle \hat{z}^i_x, z^i_y \rangle/\tau)}{\sum_{j \in [N]} \exp(\langle \hat{z}^i_x, z^j_y\rangle/\tau)}.
    \end{split}
\end{equation}

Here, $\lambda$ serves as the hyperparameter. 
$i$ and $j$ represent the index of the given input batch with the size $N$.
$\langle \cdot \rangle$ represents the inner-product and $\tau$ is the temperature parameter.
The first loss term (projection loss) measures the mean-squared error between the estimated and actual image embeddings, primarily ensuring concept alignment. 
However, our preliminary studies reveal that exclusive reliance on this term diminishes composition alignment. 
Therefore, we stick with the contrastive loss component (the second loss term) to bolster compositional generalization, with $\lambda$ balancing concept and composition alignment.

\paragraph{\textbf{Additional Control-based T2I Prior Mapping.}}
Acknowledging the limitations in existing methods, which necessitate learning the diffusion latent space even for additional control inputs, we endeavor to achieve a more nuanced balance between subject, text, and supplementary conditions.
Consequently, we have augmented \ours~to accommodate an additional modality, a Canny edge map, providing more refined control over subject-driven image generation.
This entails modifying the prior model to accept additional conditions ($f_\theta' : (z_y, z_{x_i}, z_c) \rightarrow \hat{z}_x$, here $z_c$ symbolizes the additional modality embedding). 

During training, we drop $z_c$ for 1\% to improve the unconditional generations.
This enhances the stability and broadens the generalization capabilities of \ours, yielding benefits even in the absence of these controls during inference. 
Our results demonstrate that \ours~ learns a unified mapping function, accurately estimating target image representations through the effective integration of text, image, and edge maps.

\subsection{Image-text Interleaved Training}

\begin{figure*}[!t]
\centering
    \includegraphics[width=\textwidth]{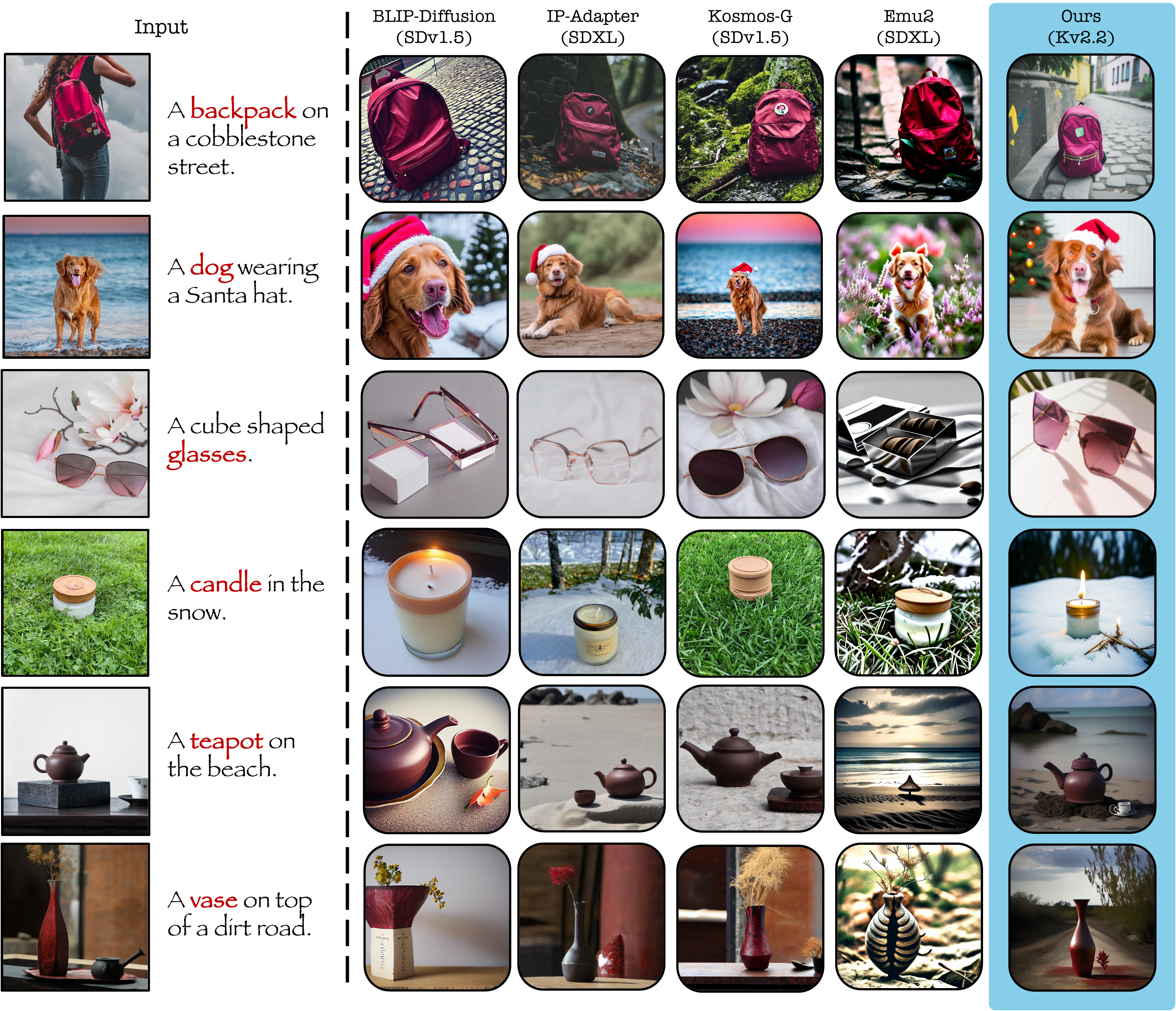}
    \caption{
    \textbf{This figure illustrates a qualitative comparison of \ours~ with contemporary approaches for single-subject T2I generations, utilizing concepts and prompts from the Dreambench dataset.} For each method, concept, and prompt, we generate four images and select the one that most accurately represents the queried concept and composition. 
    }
    \label{fig:single_subject_quantitative_main}
\end{figure*}

Our approach targets developing a versatile prior model capable of processing diverse inputs to estimate target visual outputs.
Drawing from earlier methodologies, a straightforward solution involves concatenating different inputs, like combining text (``a dog wearing sunglasses'') with respective concept-specific images. 
Preliminary experiments indicated that this method does not effectively capture the intricate relationships between target text tokens (e.g. ``dog'') and the corresponding concept images, especially when multiple concepts are present.

To address this, we adopt the interleaved pre-training strategy used in Kosmos-G, but with a notable modification to enhance resource efficiency.
We incorporate pretrained frozen CLIP text and vision encoders for extracting modality-specific embeddings—separating text-only from subject-specific images.
The key refinement in our process is the substitution of subject token-specific text embeddings with corresponding vision embeddings instead of introducing additional trainable tokens to handle the image embeddings via resampler~\cite{alayrac2022flamingo}.
First, we extract reference concept visual features ($z_{x_k} \in \mathcal{R}^{1\mathrm{x}1280}$) from the CLIP vision encoder.
Similarly, we also extract the text prompt features ($z_y \in \mathcal{R}^{77\mathrm{x}1280}$) from the last layer of the CLIP text encoder.
Here, 1280 is the CLIP-specific feature dimension.
At last, we replace the concept noun corresponding latent features from $z_y$ with $z_{x_k}$; resulting in image-text interleaved features while preserving the contextual information of the text features.
This alteration allows us to bypass the need to train the big priors models (e.g. MLLMs), significantly improving the model's proficiency in handling interleaved data.

For the generation of high-quality training datasets, we carefully selected 2 million high-quality images from the LAION dataset~\cite{schuhmann2022laion}, each with a resolution of 1024x1024. 
Utilizing BLIP2, we generate captions for these images and employ SAM~\cite{kirillov2023segment} for extracting noun or subject-specific segmentation masks.
Given the CLIP model's requirement for 224x224 resolution images, we avoid resizing the masks within their original resolutions. Instead, we opt for cropping the area of interest using Grounding DINO~\cite{liu2023grounding}, followed by resizing the masked object while preserving its aspect ratio. This technique is crucial in retaining maximum visual information for each subject during the training phase.
We provide more details about the filters used in the supplementary material.

\subsection{Additional Finetuning}
\label{sec:finetuning_main_detail}

Due to the nature of UnCLIP models, even if \ours~is very accurate, the diffusion UNet model ($h_\phi$) may not be effective in generating very unique visual representations. 
However, such behavior is common across the fast P-T2I methods and they lack in terms of maintaining performance compared to the finetuning-based methods (as outlined in Table~\ref{tab:dreambench_results}). 
Therefore, we extend the \ours~and perform concept-specific finetuning. 

Compared to the traditional finetuning methodologies (such as DreamBooth), \ours~provides very unique advantages. 
As \ours~prior model ($f_\theta$) is pre-trained for personalization, there is no need for further finetuning the $f_\theta$ and we need to only finetune diffusion UNet model.
Importantly, the fine-tuning of the $h_\phi$ does not depend on the text embeddings ($z_y$).
Hence, this leads to stable fine-tuning of the $h_\phi$; unlike DreamBooth on stable diffusion that observes catastrophic forgetting (Section~\ref{sec:appendix_finetuning}).
The new fine-tuning objective is:

\begin{equation}
    \mathcal{L}_{decoder} = 
    \underset{
        \substack{
            \epsilon \sim N(0, I) \\
            t \sim [0, T], (z_x)
            }
        }{\E}
        \Big[||\epsilon - h_{\phi}(x^{(t)},t,z_x)||_{2}^{2} \Big].
    \label{eq:finetuning}
\end{equation}

Here, $z_x$ is the visual feature of the reference concept image $x$. 
Notably, we do not need to use regularization from the DreamBooth as text alignment is already ensured during the pretraining stage of \ours. 
Moreover, this finetuning can be performed across the set of given visual concepts altogether in a single model without degrading performance. 

In summary, the prior model, trained with our image-text interleaved data and supplementary condition, presents an efficient pathway for resource-efficient multi-subject-driven image generations.

\section{Experiments}
\label{sec:experiments}

In this section, we first introduce the experimental and evaluation setups.
Later, we delve into the qualitative and quantitative results.

\paragraph{\textbf{Training and inference details.}}
We initialize our model, \ours, equipped with 34M parameters.
We train our model on an image-text interleaved dataset of 2M instances, partitioned into 1.6M for training and 0.4M for validation. 
The model is specifically tuned for the Kandinsky v2.2 diffusion image decoder.
Therefore, we use pre-trained OpenCLIP-ViT-G/14\footnote{\url{https://huggingface.co/laion/CLIP-ViT-g-14-laion2B-s12B-b42K}} as the text and vision encoders, ensuring alignment with Kandinsky v2.2 image embeddings. 
Training is executed on 2 x A100 GPUs, leveraging a per-GPU batch size of 512 and a peak learning rate of 0.00005, across approximately 100,000 iterations, summing up to 74 GPU hours.
During inference, the model employs 50 DDIM steps and 7.5 classifier-free guidance for the Kandinsky v2.2 diffusion image generator.
Adhering to baseline methodologies, we perform the P-T2I following the baseline papers' protocols.
For \ours, target subject pixel regions in reference images are segmented before embedding extraction via the CLIP(vision) encoder. 
We drop the Canny edge map during inference unless specified explicitly.
Unless specified all results (quantitative and qualitative) are without concept-specific additional fine-tuning. 

\paragraph{\textbf{Evaluation setup.}}
We primarily utilize Dreambench (encompassing 30 unique concepts with 25 prompts per concept) for qualitative and quantitative evaluations using DINO and CLIP-based metrics~\cite{ruiz2023dreambooth}.
Due to their limitations, we extend our evaluations on the ConceptBed~\cite{patel2023conceptbed} benchmark (covering 80 diverse imagenet concepts and a total of 33K composite prompts), where we report performance on concept replication, concept, and composition alignment using the Concept Confidence Deviation ($CCD$) metric~\cite{patel2023conceptbed}.
We extend Dreambench for multi-subject customization and present the Multibench dataset. 
Multibench contains about 24 unique concepts and 15 diverse prompts that result in 904 two-subject specific prompts and 1476 three-subject specific prompts. 
We provide further details about the Multibench in supplementary materials.
\begin{table}[!t]
\centering
\caption{
\textbf{Quantitative comparisons of different methodologies on Dreambench.} 
The \textbf{Bold} and \underline{underline} represent the metric-specific first and second-ranked methods, respectively. 
* represents that we re-benchmark the performance from open-source weights.
}
\label{tab:dreambench_results}
\resizebox{1\columnwidth}{!}{
    \begin{tabular}{>{\columncolor[gray]{0.95}}l|c|cc|ccc}\toprule
        \textbf{Method} &\textbf{Base Model} &\textbf{Params} &\textbf{GPU Hours} &\textbf{DINO ($\uparrow$)} &\textbf{CLIP-I ($\uparrow$)} &\textbf{CLIP-T ($\uparrow$)} \\\midrule
        \textbf{Textual Inversion} & SDv1.5 & 768 & 1 &0.569 &0.780 &{0.255} \\
        \textbf{DreamBooth} & SDv1.5 &0.9B & 0.2 &{0.668} &\underline{0.803} &\textbf{0.305} \\
        \textbf{Custom Diffusion} & SDv1.5 &57M & 0.2 &{0.643} &{0.790} &\textbf{0.305} \\
        \textbf{BLIP-Diffusion} & SDv1.5 & 0.9B & 0.1 & \underline{0.670} & \textbf{0.805} & 0.302 \\
        \hdashline
        \rowcolor{apricot!50}\textbf{\ours*} & Kv2.2 & 0.9B & 0.2 & \textbf{0.682} & 0.796 & \underline{0.304} \\
        \midrule
        \textbf{Re-Imagen} & Imagen & - & - &0.600 &0.740 &0.270 \\
        \textbf{ELITE} & SDv1.4 &77M & 336 &0.621 &0.771 &\textbf{0.293} \\
        \textbf{Subject-Diffusion} & SDv1.5 &252M &- &\textbf{0.711} &0.787 &\textbf{0.293} \\
        \textbf{BLIP-Diffusion*} &SDv1.5 &1.5B &2304 &0.603 &0.793 &0.291 \\
        \textbf{IP-Adapter*} &SDv1.5 &22M &672 &\underline{0.629} &\textbf{0.827} &0.264 \\
        \textbf{IP-Adapter*} &SDXL &22M &672 &0.613 &\underline{0.810} &0.292 \\
        \hdashline
        \textbf{Kosmos-G*} &SDv1.5 &1.9B & 12300 &\textbf{0.618} &\textbf{0.822} &0.250 \\
        \textbf{Emu2*} &SDXL &37B & - &0.563 &0.765 &\underline{0.273} \\
        \rowcolor{apricot!50}\textbf{\ours*} &Kv2.2 & \textbf{34M} & \textbf{74} &\underline{0.613} &\underline{0.783} &\textbf{0.307} \\
        \bottomrule
    \end{tabular}
}
\end{table}
\begin{table}[!t]
\centering
\caption{
\textbf{Quantitative comparisons of different methodologies on ConceptBed.} 
We present results on $CCD~(\downarrow)$ across three evaluation categories.
The \textbf{Bold} and \underline{underline} represent the metric-specific first and second-ranked methods, respectively. 
* represents our benchmarking.
}
\label{tab:conceptbed_results}
    \begin{tabular}{>{\columncolor[gray]{0.95}}l|c|ccc}\toprule
        \textbf{Method} &\textbf{Base} &\textbf{Concept} &\textbf{Concept} &\textbf{Composition} \\
        & \textbf{Model} & \textbf{Replication ($\downarrow$)} & \textbf{Alignment ($\downarrow$)} & \textbf{Alignment ($\downarrow$)}\\
        \midrule
        \textbf{Textual Inversion} &SDv1.4 &\textbf{0.0662} &\textbf{0.1163} &0.1436 \\
        \textbf{Dreambooth} &SDv1.4 &\underline{0.0880} &\underline{0.3551} &\underline{0.0360} \\
        \textbf{Custom Diffusion} &SDv1.4 &0.2309 &0.4882 &\textbf{0.0204} \\
        \midrule
        \textbf{ELITE*} &SDv1.4 &\underline{0.3195} &0.4666 &0.1832 \\
        \textbf{BLIP-DIffusion*} &SDv1.5 & 0.3510&\textbf{0.3245} &0.1589 \\
        \textbf{IP-Adapter*} &SDXL & 0.3665 &\underline{0.3571} &\underline{0.0641} \\
        \hdashline
        \rowcolor{apricot!50}\textbf{\ours*} &Kv2.2 & \textbf{0.2853} & 0.3619 & \textbf{-0.0200} \\
        \bottomrule
    \end{tabular}
\end{table}

\subsection{Results \& Analysis}

\begin{figure*}
    \centering
    \addtolength{\leftskip} {-4cm}
    \addtolength{\rightskip}{-4cm}
    \includegraphics[width=1.2\textwidth]{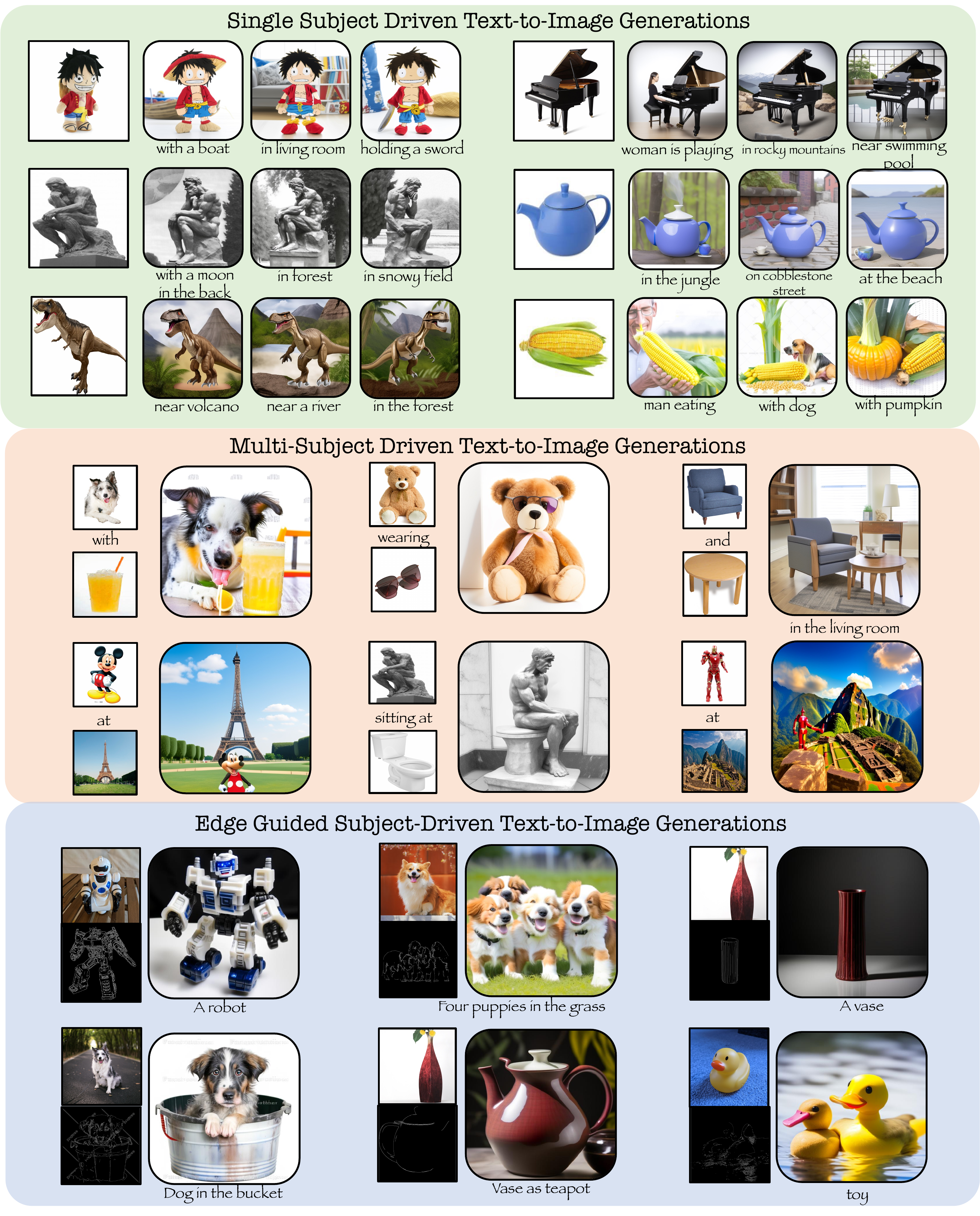}
    \caption{
     Qualitative results categorized by generative capabilities.
    }
    \label{fig:full-page-examples}
\end{figure*}

\paragraph{\textbf{Quantitative comparison.}}
The quantitative assessments detailed in Table~\ref{tab:dreambench_results} and Table~\ref{tab:conceptbed_results} focus on the single-concept T2I task, while Table~\ref{tab:multibench_results} shows the results on multi-concept-driven image generation.
For Dreambench and Multibench, we generate and evaluate four images per prompt, reporting average performance on three metrics (DINO, CLIP-I, and CLIP-T). 
In the case of ConceptBed, we process each of the 33K prompts to generate a single concept image.
The results, as depicted in these tables, highlight \ours's superior performance in composition alignment while maintaining competitive concept alignment.
Analysis on ConceptBed (Table~\ref{tab:conceptbed_results}) indicates that \ours~exhibits a notable proficiency in concept replication, albeit with a marginal trade-off in concept alignment for enhanced composition fidelity. 
Comparatively, all baselines prioritize concept alignment, often at the expense of composition alignment.
While \ours~improves the CLIP-T while preserving the DINO; achieved with significantly fewer resources.

However, a significant gap remains between finetuning-based and fast P-T2I methods.
We further perform concept-specific fine-tuning (as described in Section~\ref{sec:finetuning_main_detail}).
As shown in Table~\ref{tab:dreambench_results}, \ours~outperforms the DreamBooth and BLIP-Diffusion in terms of concept alignment (DINO) while maintaining the performance on composition alignment (CLIP-T). 
Notably, Multibench results (Table~\ref{tab:multibench_results}) indicate that \ours~significantly outperforms the Kosmos-G (2B params) and Emu2 (37B params) in terms of CLIP-T while maintaining the DINO performance.
Therefore, we can conclude that \ours~is the most resource-efficient compared, especially when compared to the MLLM-based methods.

\begin{table}[!t]
\centering
\caption{
\textbf{Quantitative comparisons of different methodologies on Multibench.} 
The \textbf{Bold} and \underline{underline} represent the metric-specific first and second-ranked methods on each metric, respectively. 
}

\label{tab:multibench_results}
    \begin{tabular}{l|ccc|cc}\toprule
        & \multicolumn{3}{c|}{\textbf{Two Subjects}} & \multicolumn{2}{c}{\textbf{Three Subjects}} \\
        & \textbf{Kosmos-G} & \textbf{Emu2} & \cellcolor{apricot!50}\textbf{\ours} & \textbf{Emu2} & \cellcolor{apricot!50}\textbf{\ours} \\
        \midrule
        \textbf{DINO ($\uparrow$)} & \textbf{0.4549} & 0.4451 & \cellcolor{apricot!50}\underline{0.4478} & 0.3168 & \cellcolor{apricot!50}\textbf{0.3420}\\
        \textbf{CLIP-I ($\uparrow$)} & \textbf{0.7759} & 0.7397 & \cellcolor{apricot!50}\underline{0.7409} & 0.6231 & \cellcolor{apricot!50}\textbf{0.6463} \\
        \textbf{CLIP-T ($\uparrow$)} & 0.2493 & \underline{0.2673} & \cellcolor{apricot!50}\textbf{0.3327} & 0.2819 & \cellcolor{apricot!50}\textbf{0.3469} \\
        \bottomrule
    \end{tabular}
\end{table}
\begin{table}[!t]
\centering
\caption{
\textbf{Quantitative results of Canny edge controlled P-T2I of different methodologies on Dreambench.} 
The \textbf{Bold} and \underline{underline} represent the metric-specific first and second-ranked methods, respectively. 
Red highlighted numbers indicate the relative percentage drop for concept alignment compared to Table~\ref{tab:dreambench_results}.
}
\label{tab:edge_results}
    \begin{tabular}{>{\columncolor[gray]{0.95}}l|ccc}\toprule
        \textbf{Method} &\textbf{DINO ($\uparrow$)} &\textbf{CLIP-I ($\uparrow$)} &\textbf{CLIP-T ($\uparrow$)} \\\midrule
        \textbf{BLIP-Diffusion*} &0.4234\nresults{29.7\%} &0.7119 &\underline{0.3152}\\
        \textbf{IP-Adapter*} &\underline{0.4281}\nresults{31.9\%} &\underline{0.7315} &0.3034 \\
        \hdashline
        \rowcolor{apricot!50}\textbf{\ours*} &\textbf{0.5173}\nresults{14.3\%} &\textbf{0.7437} &\textbf{0.3158} \\
        \bottomrule
    \end{tabular}
\end{table}

\begin{figure}
\centering
\begin{subfigure}{0.85\textwidth}
  \centering
  \includegraphics[width=\linewidth]{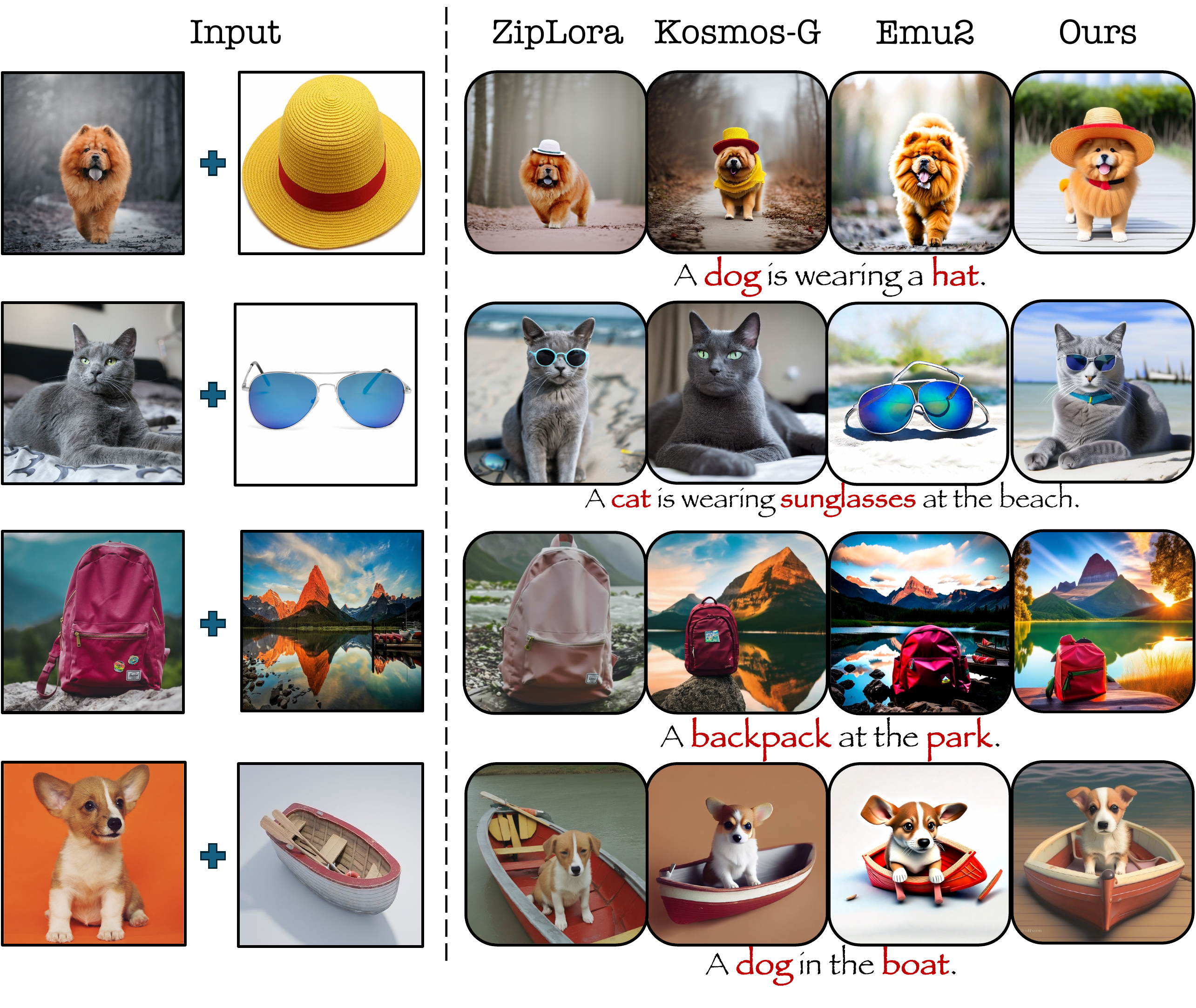}
    \caption{
    \textbf{Multi-subject qualitative examples}.
    }
    \label{fig:multi-subject-examples}
\end{subfigure}
\begin{subfigure}{0.8\textwidth}
  \centering
  \includegraphics[width=\linewidth]{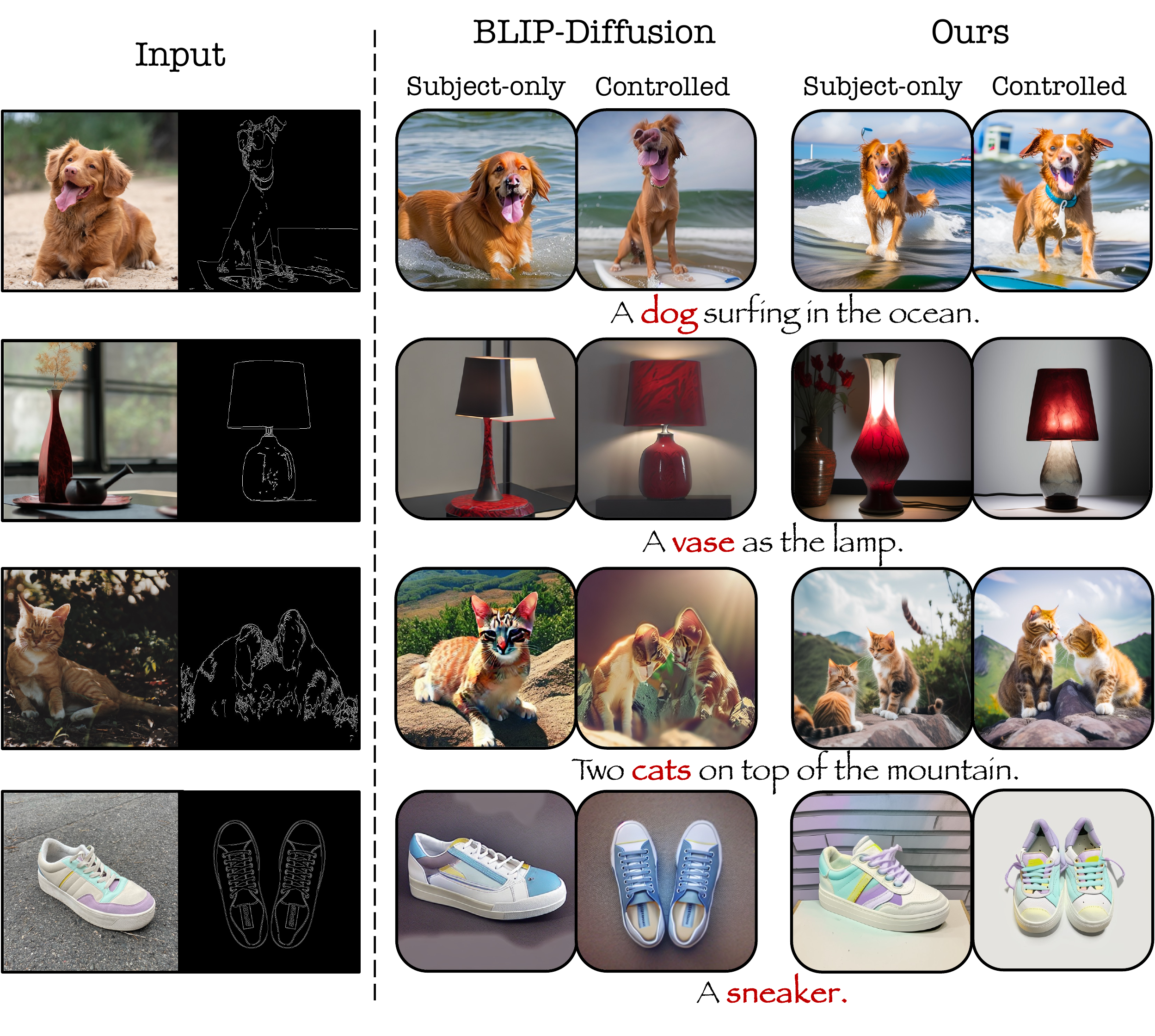}
    \caption{
    \textbf{Qualitative examples for edge-guided P-T2I}.
    }
    \label{fig:control-net-examples}
\end{subfigure}
\caption{
    \textbf{Qualitative comparison} between \ours~and other baselines.
    }
\end{figure}

\paragraph{\textbf{Qualitative comparisons.}}
In Figure~\ref{fig:single_subject_quantitative_main}, we present a range of single subject-specific images generated by various methodologies including BLIP-Diffusion, IP-Adapter, Kosmos-G, Emu2, and \ours.
\ours~demonstrates exemplary proficiency in composition while ensuring concept alignment.
In contrast, the baselines often overemphasize reference images or exhibit concept dilution, leading to higher concept alignment but compromised composition.
Interestingly, we find that Emu2 can capture the single-subjects but it fails to reproduce them with complex text compositions (as shown in Figure~\ref{fig:single_subject_quantitative_main}).
Similarly, Figure~\ref{fig:multi-subject-examples} exhibits \ours's multi-concept generation prowess, in comparison to ZipLoRA (fine-tuning-based approach) along with Kosmos-G and Emu2 (Multimodal LLM-based approaches), underscoring its capability to rival compute-intensive methods.
We discuss additional examples and limitations in the appendix.
That said, even though \ours~improves the performance over the baselines, this is still not enough and it signifies the challenges associated with fast multi-concept personalization.

\paragraph{\textbf{Canny edge controlled image generation.}}
As shown in Figure~\ref{fig:control-net-examples}, the baseline (BLIP-Diffusion) adheres strictly to the imposed edge maps, often at the cost of concept retention (rows 1, 3, and 4).
This leads to a large number of unwanted artifacts in the generated images.
To further ground this behavior, we first generated images using Stable Diffusion v1.5 for Dreambench prompts without customization then we extracted the Canny edge map and used this edge map to control the subject-driven image generations.
At last, we report the performance in Table~\ref{tab:edge_results}. 
It can be observed that both baselines IP-Adapter and BLIP-Diffusion drop the DINO score by 30\%, which follows the qualitative results.
While \ours~do not follow the Canny edge precisely but preserves the concept alignment and improves the performance relatively by 21\%.

\begin{table}[!t]
\centering
\caption{\textbf{Ablation studies} w.r.t. to the key components of \ours~ design. We report the concept and composition alignment for single-subject T2I using $CCD~(\downarrow)$ on the ConceptBed benchmark.}
\label{tab:ablation}
    \begin{tabular}{>{\columncolor[gray]{0.95}}l|cc}\toprule
        \textbf{Model} &\textbf{Concept} &\textbf{Composition} \\
        & \textbf{Alignment ($\downarrow$)} & \textbf{Alignment ($\downarrow$)}\\
        \midrule
        Projection loss (i.e. $\lambda$=0.0) &0.394 &0.008 \\
        \quad w/ contrastive loss ($\lambda$=0.5) &0.435 &\textbf{-0.043} \\
        \quad w/ contrastive loss ($\lambda$=0.2) &0.402 &-0.026 \\
        \hdashline
        \quad w/ edge conditions ($\lambda$=0.2) &\textbf{0.362} &-0.020 \\
        \bottomrule
    \end{tabular}
\end{table}

\begin{figure}[!t]
\centering
    \includegraphics[width=0.7\linewidth]{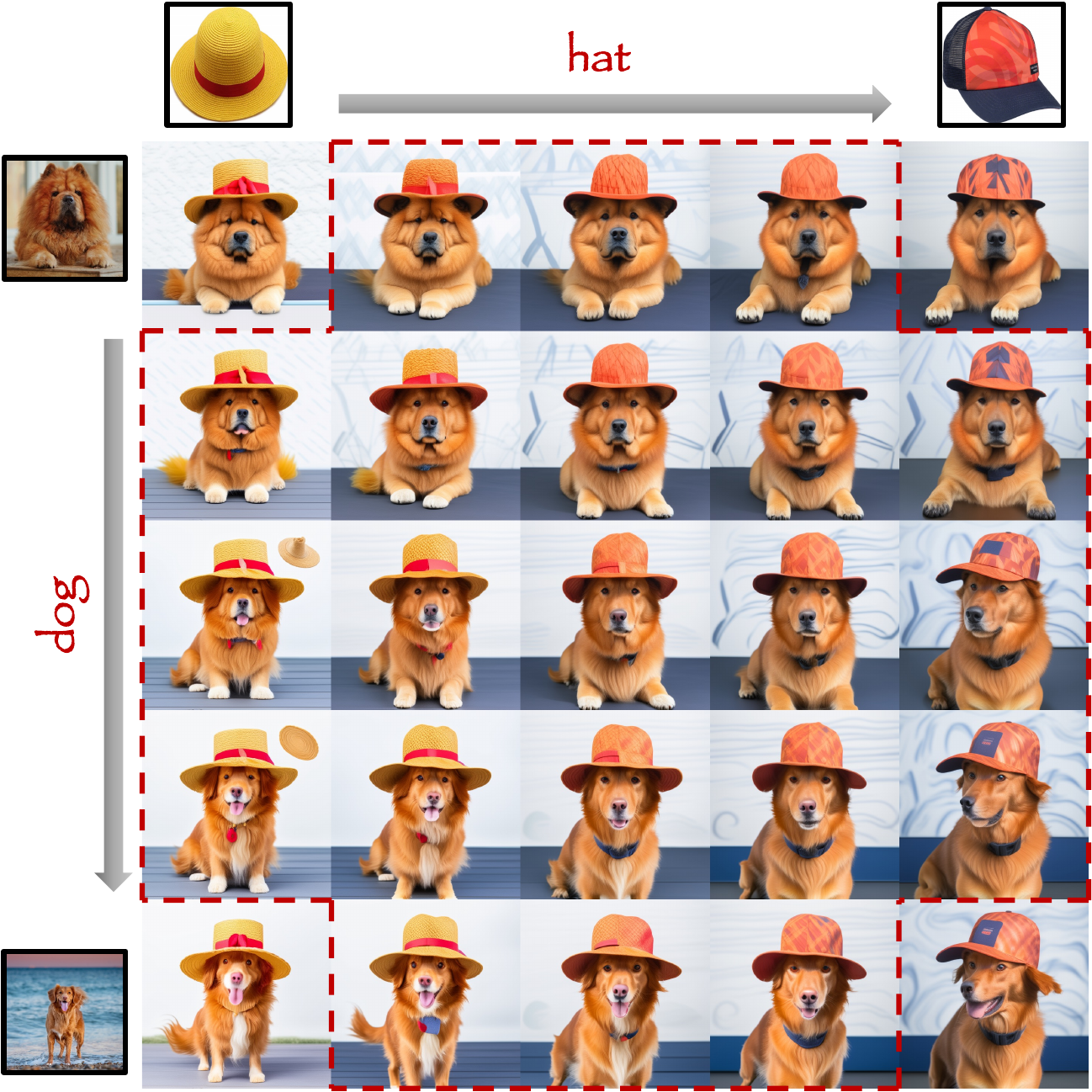}
    \caption{
    \textbf{Interpolation between four concepts.} Here, we estimate the image embedding using \ours~corresponding to each corner and then interpolate from top to bottom and left to right.
    At last, we use the Kandinsky v2.2 diffusion UNet model to generate the images with fixed random seeds from these sets of image embeddings.
    }
    \label{fig:interpolation}
\end{figure}

\paragraph{\textbf{Ablations.}}
We extend our study to evaluate the individual contributions of different components in \ours. 
Initially, the model's performance with solely the projection loss (referenced in Eq.\ref{eq:eclipse}) is assessed.
Subsequent experiments involve training \ours~variants with varying hyperparameters for the contrastive loss, specifically $\lambda$ values of 0.2 and 0.5.
A comparative analysis of these baselines is conducted against the fully equipped \ours~model, which incorporates $\mathcal{L}_{prior}$ (Eq.\ref{eq:eclipse}) with $\lambda=0.2$ and utilizes Canny edge maps during training.
Relying solely on projection loss results in high concept alignment but compromises compositions (Table~\ref{tab:ablation}).
The contrastive loss variant with $\lambda=0.5$ enhances composition alignment at the expense of concept alignment, whereas $\lambda=0.2$ achieves a more balanced performance.
Crucially, the integration of Canny edge maps during training optimally balances both alignments and, specifically, improves the concept alignment.
The negative values indicate that the $CCD$ oracle model is highly confident in the generated images.

\paragraph{\textbf{Multi-subject interpolation.}}
A key attribute of the CLIP latent space is the ability to perform smooth interpolation between two sets of embeddings. 
We conducted experiments to demonstrate \ours's ability to learn and replicate this smooth latent space transition.
We selected two distinct dog breeds ($<$dog1$>$, $<$dog2$>$) and two types of hats ($<$hat1$>$, $<$hat2$>$) as the concepts.
\ours~was then used to estimate image embeddings for all four possible combinations, each corresponding to the input phrase ``a $<$dog$>$ wearing a $<$hat$>$.'' 
Fig.~\ref{fig:interpolation} showcases a gradual and seamless transition in the synthesized images from the top left to the bottom right.
Unlike current diffusion models, which often exhibit sensitivity to input variations requiring numerous iterations of user interactions for desired outcomes, \ours~inherits CLIP's smooth latent space.
This not only facilitates progressive changes in the conceptual domain but also extends the model's utility by enabling personalized \textbf{multi-subject interpolations}.

\section{Conclusion}
In this paper, we have introduced a novel diffusion-free methodology for personalized text-to-image (P-T2I) applications, utilizing the latent space of the pre-trained CLIP model with high efficiency.
Our \ours~model, trained on an image-text interleaved dataset, achieves the capability to execute single-concept, multi-concept, and edge-guided controlled P-T2I tasks using a singular model framework, while simultaneously minimizing resource utilization.
Notably, \ours~sets a new benchmark in achieving competitive performance in terms of concept and composition alignment.
Furthermore, our research illuminates the potential of \ours~in exploring and leveraging the smooth latent space. This capability unlocks new avenues for interpolating between multiple concepts and their amalgamation, thereby generating entirely novel concepts.
Our findings underscore a promising pathway to improve MLLMs to effectively control the pre-trained diffusion image models without necessitating extra supervision. 

\section*{Acknowledgments}
This work was supported by NSF RI grants \#1750082, \#2132724, and CPS grant \#2038666. 
We thank the Research Computing (RC) at Arizona State University (ASU) for providing computing resources.
The views and opinions of the authors expressed herein do not necessarily state or reflect those of the funding agencies and employers.

\setcitestyle{numbers,square}
\bibliographystyle{plain}
\bibliography{main}

\clearpage
\appendix

\section{Preliminaries for T2I Diffusion Models}
As evidenced in numerous contemporary studies regarding T2I models, Stable Diffusion (SD)~\cite{rombach2022high} has emerged as a predominant backbone for T2I models. 
SD involves training diffusion models in latent space, reversing a forward diffusion process that introduces noise into the image. 
A notable feature of SD is its integration of cross-attention, facilitating various conditions like text input. 
Operating in VQ-VAE latent space, SD not only achieves exceptional generative performance surpassing that in pixel space but also significantly reduces the computational demands.

UnCLIP models (such as DALL-E 2) are very similar to the Stable Diffusion.
In contrast, the UnCLIP takes the modular approach.
UnCLIP first trains the diffusion text-to-image to the image prior ($f_{\theta}$) to estimate the image embeddings ($z_x$) from the text embeddings ($z_y$).
In parallel, a UNet-like diffusion image generator ($h_{\phi}$) is trained to generate images ($x$) conditioned on ground truth vision embeddings ($z_x$). 

Traditionally, T2I prior is modeled to estimate $x_0$-prediction instead of $\epsilon$-prediction.
Given the forward function $z_x^{(t)} \sim q(t, z_x)$, the goal of $f_\theta$ is to directly estimate $z_x$ for all timesteps $t \sim [0, T]$ as:
\begin{equation}
    \mathcal{L}_{prior} = 
    \underset{
        \substack{
            t \sim [0, T], \\
            z_x^{(t)} \sim q(t, z_x)
            }
        }{\E}
        \Big[||z_x - f_{\theta}(z_x^{(t)},t,z_y)||_{2}^{2} \Big].
    \label{eq:prior}
\end{equation}

\eclipse~proposes the contrastive learning strategy (Eq.~1 -- main paper) instead of minimizing Eq.~\ref{eq:prior}. 
The diffusion image generator is trained by following standard $\epsilon$-prediction formulation. 
Here, $h_\phi$ will estimate the ground truth added Gaussian noise $\epsilon \sim N(0,I)$, given the noise image $X^{(t)}$ for all timesteps $t \sim [0,T]$ and input conditions (such as $z_x$, $z_y$).

\begin{equation}
    \mathcal{L}_{decoder} = 
    \underset{
        \substack{
            \epsilon \sim N(0, I) \\
            t \sim [0, T], \\
            (z_x,~z_y)
            }
        }{\E}
        \Big[||\epsilon - h_{\phi}(x^{(t)},t,z_x,z_y)||_{2}^{2} \Big].
    \label{eq:decoder}
\end{equation}

For models like Kandinsky v2.2, we drop the $z_y$ to condition the model on $z_x$.
Therefore, \ours~also only conditions the image generation with $z_y$ in the prior stage.

\section{Image-Text Interleaved Training Details}
\paragraph{\textbf{Dataset Creation}}

\begin{figure}[!t]
\centering
    \includegraphics[width=0.7\linewidth]{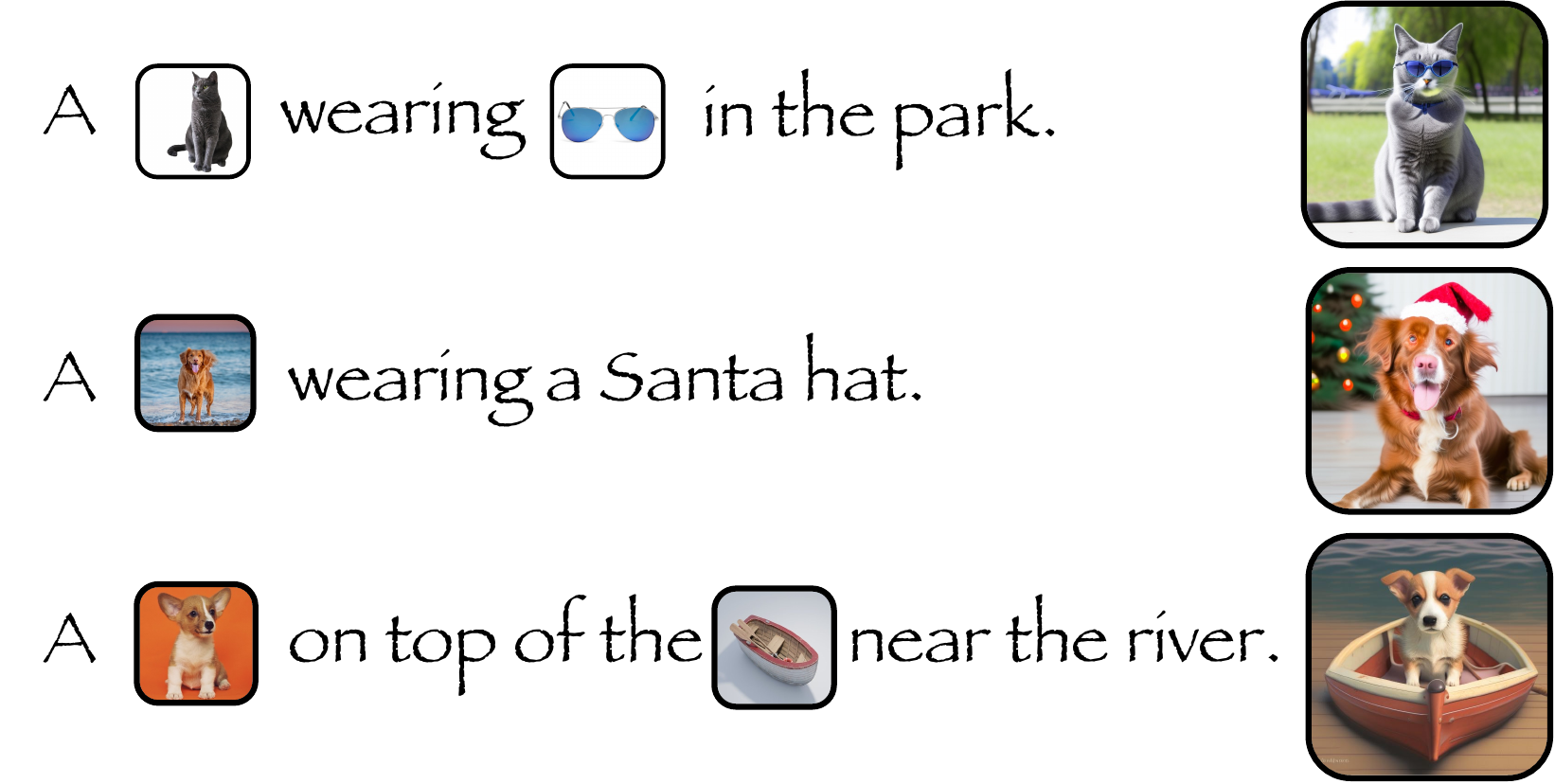}
    \caption{
     \textbf{Examples of image-text interleaved training data.}
     The left column shows the input of the prior model and the right images shows the ground truth corresponding images.
     Note: these examples are generated from \ours~for better interpretability.
    }
    \label{fig:interleaved}
\end{figure}

In constructing the dataset, we adhered to the data processing pipeline of Subject Diffusion~\cite{ma2023subject}.
We utilized the LAION-5B High-Res dataset, requiring a minimum image size of 1024x1024 resolution. 
Original captions were replaced with new captions generated by BLIP-2 (flan-t5-xl)\footnote{\url{https://huggingface.co/Salesforce/blip2-flan-t5-xl}}. 
Subjects were extracted using Spacy\footnote{\url{https://spacy.io}}. 
For each subject, we identified bounding boxes employing Grounding DINO~\cite{liu2023grounding}, setting both box-threshold and text-threshold values to 0.2. 
We retained images with 1 to 8 detected bounding boxes, discarding the rest. 
Additionally, captions with multiple instances of identical objects were filtered, allowing a maximum of 6 identical objects. 
Following bounding box detection, individual subject masks were isolated using Segment-Anything (SAM)~\cite{kirillov2023segment}.
To enhance the efficiency of the training process, we pre-processed the dataset by pre-extracting features from CLIP vision and text encoders.
During this phase, images predominantly featuring a background (white portion) exceeding 10\% of the total area were excluded.  
We preserved bounding boxes with a width-height ratio ranging from 0.08 to 0.7 and logit scores of at least 0.3. 
Masks constituting less than 40\% of the bounding box area were discarded. 
For the selection of subjects in images, we constrained the range to 1-4 subjects per image, excluding those with more than 4 subjects.
At last, the interleaved image-text examples with respective ground truth images are shown in Figure~\ref{fig:interleaved}.

\paragraph{\textbf{Dataset Statistics}}

In the final analysis, our dataset comprised a total of 1,990,123 images. The distribution of subjects per image exhibited a range from 1 to 4, with the following breakdown: 1,479,785 images featuring one subject, 432,831 images with two subjects, 65,597 images containing three subjects, and 11,910 images showcasing four subjects. The overall count of unique subjects acquired from this dataset amounted to 30,358.
We partitioned our dataset into an 80:20 split between training and validation, reserving the remaining 1.6 million images for training and the rest for validation.

\section{Implementation Details}
The \ours~transformer prior architecture is significantly more compact compared to other Text-to-Image (T2I) methodologies. 
Our model employs a configuration of 16 Attention Heads, each with a dimension size of 32, alongside a total of 10 layers. 
Additionally, the embedding dimension size for our model is set at 1280, supplemented by 4 auxiliary embeddings (including, one for canny edge map).
As \ours~is not a diffusion prior model, we do not keep time embedding layers.
Overall, the \ours~model comprises approximately 34 million parameters, establishing it as a streamlined yet effective solution for Personalaized-T2I.
Notably, the standard UnCLIP T2I priors contain 1 billion parameters.

\section{\ours~with Finetuning}
\label{sec:appendix_finetuning}
\begin{figure*}[!t]
    \includegraphics[width=\textwidth]{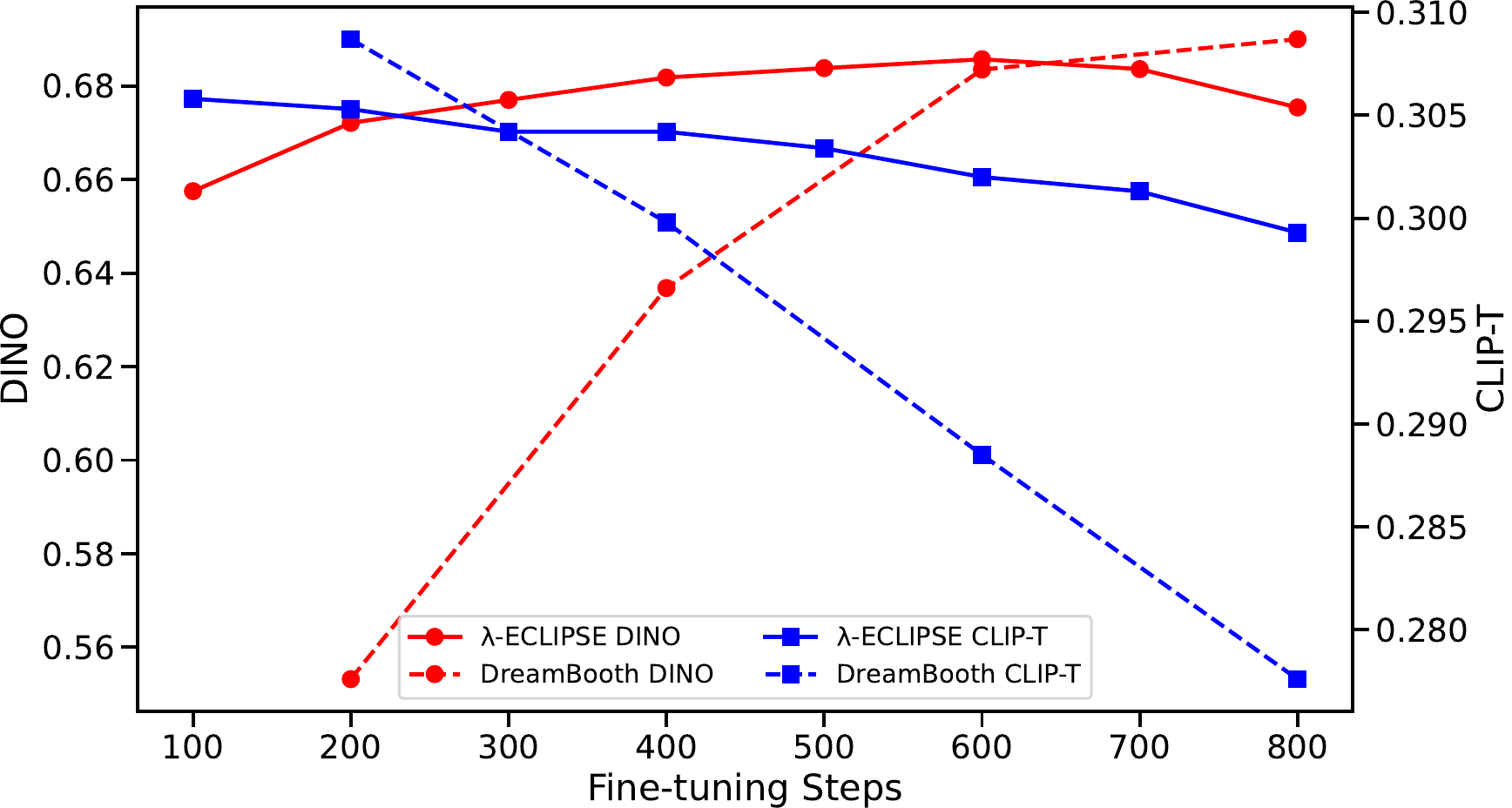}
    \caption{
     \textbf{DreamBooth (Stable Diffusion v1.5) \textit{vs.} \ours~ (with fine-tuning)} \textit{w.r.t.} DINO and CLIP-T metrics on Dreambench.
    }
    \label{fig:finetuning}
\end{figure*}

As demonstrated in the main paper (Table~\ref{tab:dreambench_results}), the superiority of fine-tuning-based personalization methodologies, whether applied to single-subject or multi-subject frameworks, over non-fine-tuning alternatives is evident. 
Consequently, we have expanded our analysis through additional fine-tuning of the \ours.

\paragraph{\textbf{Experimental Setup.}}
Given that \ours~effectively trains the T2I prior, capturing concept-specific features to a significant degree, we opted not to further optimize this component. 
Our focus shifted to exclusively fine-tuning the diffusion UNet model ($h_\phi$), employing the AdamW optimizer at a learning rate of 1e-5, without warm-up steps. 
For the DreamBooth application within the Stable Diffusion v1.5 model, we selected a learning rate of 5e-6, maintaining consistency in other hyperparameters. 
To simplify, we excluded the use of a prior preservation regularizer and conducted training on the Dreambench platform using a single RTX A6000 GPU.

\paragraph{\textbf{Results.}}
Our findings, illustrated in Figure~\ref{fig:finetuning}, reveal that \ours~and DreamBooth exhibit improved performance with incremental fine-tuning steps. 
Notably, the DINO score improved from 0.61 to 0.68 with few optimization steps and outperforms the baselines (see Table~\ref{tab:dreambench_results}). 
A detailed analysis indicates that while DreamBooth's DINO score improves, its CLIP-T performance diminishes, hinting at concept overfitting. 
Conversely, \ours~consistently improves in DINO scoring without adversely impacting the CLIP-T performance, underscoring the efficacy of our image-text interleaved training approach at the prior stage. 
Qualitative comparisons, as shown in Figure~\ref{fig:finetuning-examples}, further highlight the benefits of fine-tuning \ours~with minimal steps.

\paragraph{\textbf{Advantages of fine-tuning \ours.}}
The fine-tuning of \ours, in comparison to the baselines, reveals two key benefits:
1) Achieving state-of-the-art (SOTA) performance within a few finetuning steps.
2) Unlike the Stable Diffusion model, which exhibits catastrophic forgetting of nearby concepts post-DreamBooth fine-tuning, \ours~maintains previous knowledge. 
This suggests that a single model is sufficient to effectively fine-tune across multiple concepts together.

This analysis underscores the strategic advantages and enhanced efficiency of fine-tuning \ours~for personalized applications in complex visual data processing.

\section{Extended P-T2I Baselines Comparison}

\begin{table*}[!t]
\centering
\small
\caption{
\textbf{The detailed overview of subject-driven text-to-image generative methodologies.}
* represents the backbone base models listed are subject to potential updates or modifications.
} 
\setlength{\tabcolsep}{8pt}
\resizebox{\textwidth}{!}{
\begin{tabular}{r|cccc}
  \hline
\toprule
\textbf{Method} &\textbf{Multi-Subject} &\textbf{Finetuning-Free} &\textbf{Base-Model} &\textbf{\# of Input Images} \\\midrule
Re-Imagen~\cite{chen2022re} &\xmark &\cmark & Imagen &Single \\
Textual Inversion~\cite{gal2022image} &\xmark &\xmark &SDv1.4 &Multiple \\
DreamBooth~\cite{ruiz2023dreambooth} &\xmark &\xmark &SDv1.4 &Multiple \\
Custom Diffusion~\cite{kumari2023multi} &\cmark &\xmark &SDv1.4 &Multiple \\
ELITE~\cite{Wei2023ELITEEV} &\xmark &\cmark &SDv1.4 &Single \\
E4T~\cite{gal2023encoder} &\xmark &\xmark &SD &Single \\
Cones~\cite{liu2023cones} &\cmark &\xmark &SDv1.4 &Single \\
SVDiff~\cite{han2023svdiff} &\cmark &\xmark &SD &Multiple \\
UMM-Diffusion~\cite{ma2023unified} &\xmark &\cmark &SDv1.5 &Single \\
XTI~\cite{voynov2023p+} &\xmark &\xmark &SDv1.4 &Multiple \\
Continual Diffusion~\cite{smith2023continual} &\cmark &\xmark & - &Multiple \\
InstantBooth~\cite{shi2023instantbooth} &\xmark &\cmark &SDv1.4 &Multiple \\
SuTi~\cite{chen2023subject} &\xmark &\cmark &Imagen &Multiple \\
Taming~\cite{jia2023taming} &\xmark &\cmark &Imagen &Single \\
BLIP-Diffusion~\cite{li2023blip} &\xmark &\cmark &SDv1.5 &Single \\
Cones 2~\cite{liu2023customizable} &\cmark &\xmark &SDv2.1 &Single \\
DisenBooth~\cite{chen2023disenbooth} &\xmark &\xmark &SDv2.1 &Single \\
\rowcolor{apricot!50} 
 FastComposer~\cite{xiao2023fastcomposer} &\cmark &\cmark &SDv1.5 &Single \\
Perfusion~\cite{tewel2023key} &\cmark &\xmark &SDv1.5 &Multiple \\
Mix-of-Show~\cite{gu2023mix} &\cmark &\xmark &Chilloutmix &Multiple \\
NeTI~\cite{alaluf2023neural} &\xmark &\xmark &SDv1.4 &Mulitple \\
Break-A-Scene~\cite{avrahami2023break} &\cmark &\xmark &SDv2.1 &Single\textsuperscript{*} \\
ViCo~\cite{tumanyan2023plug} &\xmark &\xmark &SDv1.4 &Mulitple \\
Domain-Agnostic~\cite{arar2023domain} &\xmark &\xmark & - &Single \\
\rowcolor{apricot!50} Subject-Diffusion~\cite{ma2023subject} &\cmark &\cmark &SDv2 &Single \\
HyperDreamBooth~\cite{ruiz2023hyperdreambooth} &\xmark &\xmark &SDv1.5 &Single \\
IP-Adapter~\cite{ye2023ip} &\xmark &\cmark &SDv1.5 &Single \\
\rowcolor{apricot!50} Kosmos-G~\cite{pan2023kosmos} &\cmark &\cmark &SDv1.5 &Single \\
Zip-LoRA~\cite{shah2023ziplora} &\cmark &\xmark &SDXL &Multiple \\
CatVersion~\cite{zhao2023catversion} &\xmark &\xmark &SDv1.5 &Multiple \\
\rowcolor{apricot!50}  SSR-Encoder~\cite{zhang2023ssr} &\cmark &\cmark &SDv1.5 &Single \\
\rowcolor{apricot!50}  Emu2~\cite{sun2023generative} & \cmark & \cmark & SDXL & Single \\
\hdashline
\rowcolor{apricot!50}  \textbf{\ours~(ours)} &\cmark &\cmark &Kv2.2 &Single \\
   \hline
\end{tabular}

}
\label{tab:all_methods}
\end{table*}

We further expand our comparative analysis of P-T2I methods encompassing a total of 33 approaches including ours and parallel works.
Table~\ref{tab:all_methods} summarizes them into four crucial aspects:
1) multi-subject support, 
2) fine-tuning free,
3) base model types, and 
4) the required number of input images.
To summarize, \ours~is the only methodology built on top of the UnCLIP models while supporting multi-subject driven image generation with fine-tuning free, and only requires a single reference image input for the training.
We detail the comparison below:

\paragraph{\textbf{Multi-Subject Generation.}}
Multi-subject generation enables users to integrate multiple personal subjects to generate an image that follows the text prompts and aligns with all the concept visuals. In total, 15 of the 33 methods offer this capability, while 6 methods support fast multi-subject personalization, others demand separate training for each subject to be learned and then an additional fusing step for combining the learned subjects is required (i.e. Zip-LoRA, Mix-of-Show). Among these methods, only a few can learn auxiliary guided information such as canny edge, depth maps, or open-pose and adapt style variation (i.e. Kosmos-G).

\paragraph{\textbf{Fine-tuning Free (Fast Personalization).}}
Many methods require test-time fine-tuning. 
Each varies on which part alteration occurs, as early models tend to modify the whole UNet. 
In contrast, recent models tune a small portion of the cross-attention layers or introduce additional layers performing as adapters. 
In our analysis of P-T2I methodologies, 14 out of 33 methods employ a finetuning-free approach which enables fast personalization. 

\paragraph{\textbf{Diffusion Independent.}}
A majority of the reviewed models utilize diffusion models, with Stable Diffusion being the predominant choice, spanning versions 1.4, 1.5, 2.1, and XL. 
Few adapt Imagen (SuTi, Taming) and Mix-of-show employs ChillOutMix as their pre-trained model, known for its adeptness at preserving realistic concepts like human faces. 
A unique outlier in this landscape is our \ours, the only one that eschews the use of any diffusion prior model.

\paragraph{\textbf{Easiness of Use.}}
A more user-friendly model typically requires a single reference image per subject, as opposed to multiple images of the same subject. 
In our study, 19 methods offer P-T2I capabilities with just one input image. 
In contrast, others often require 4 to 5 images of the subject. 
Additionally, some methods necessitate storage space for learned concepts, ranging from a few hundred kilobytes (e.g., Perfusion, HyperDreamBooth) to several megabytes (e.g., Zip-LoRA). 
Our method stands out by eliminating the need for individual concept pre-learning or storing any artifacts for P-T2I utilization, offering a streamlined, efficient user experience.

\section{Multibench Dataset}

\begin{table*}[!t]
\centering
\caption{\textbf{Example of prompt templates used for Multibench dataset.} Subjects presented in Table~\ref{tab:multibench_dataset_details} are placed in \{\}.}
\setlength{\tabcolsep}{8pt}
\begin{tabular}{l|l}
\toprule
Two subjects & Three subjects \\ \midrule    
\{\} in the \{\} & \{\} with a \{\} and \{\} \\
\{\} wearing a \{\} & \{\} is playing with \{\} in \{\} \\
\{\} chasing a \{\} & \{\} with \{\} in front of \{\} \\
\{\} looking at a \{\} & \{\} with a \{\} and a view of the \{\} \\
\{\} is sitting on a \{\} & \{\} with a \{\} and \{\} in the background \\
\{\} standing on a \{\} & \\
\{\} and \{\} playing in the garden & \\
\{\} and \{\} on top of the mountain & \\
\{\} and \{\} in the jungle & \\
\{\} and \{\} in the snow & \\
\{\} and \{\} on the beach & \\
\{\} and \{\} on a cobblestone street & \\
\{\} and \{\} standing next to each other & \\
\bottomrule
\end{tabular}
\label{tab:prompts}
\end{table*}

\begin{table*}[!]
\centering
\caption{\textbf{Number of occurrences of unique subject categories.} The left side of the table are subjects used for two subjects prompts, and the right side of the table are subjects used for three subjects prompts.}
\begin{tabular}{lr|lr|lr|lr}
\toprule
\multicolumn{4}{c|}{Two subjects}                         & \multicolumn{4}{c}{Three subjects}                   \\ \midrule
dog        & 76 & boat        & 5 & dog           & 81 & rainbow        & 35 \\
cat        & 76 & park        & 4 & stuffed animal & 105 & ruins          & 35 \\
bird       & 76 & ruins       & 9 & toy           & 105 & tower          & 35 \\
horse      & 73 & castle      & 5 & cat           & 81 & horse          & 81 \\
guinea pig & 73 & desert      & 4 & desert        & 60 & bird           & 81 \\
glasses    & 5  & rainbow     & 5 & hill          & 60 & guinea pig     & 81 \\
hat        & 5  & candle      & 5 & castle        & 45 & guitar         & 25  \\
tower      & 10  & backpack    & 3  & backpack      & 65 & french horn    & 25  \\
           &     &             &    & can           & 130 & vase           & 25  \\
           &     &             &    & candle        & 65 & robot          & 25  \\
           &     &             &    & church        & 35 &                &     \\
\bottomrule
\end{tabular}
\label{tab:multibench_dataset_details}
\end{table*}

We provide additional qualitative results in Figure~\ref{fig:full-page-examples}.
For the multi-subject image benchmark, our dataset comprises 2,308 unique prompts, segmented into 904 two-subject and 1,476 three-subject prompts. 
This dataset integrates subjects from the original DreamBench dataset, featuring 30 distinct concepts. 
We expanded the dataset by incorporating additional concepts vital for two and three subject-specific prompts, such as various parks, hats, glasses, and more. 
Prompt templates and the count of unique subject categories featured in prompts are detailed in Tables~\ref{tab:prompts} and ~\ref{tab:multibench_dataset_details}, respectively. 
Overall, the dataset includes 217 two-subject compositions and 405 three-subject compositions, enriching the benchmark's diversity and comprehensiveness.

\section{Qualitative Results \& Failure Cases}
\begin{figure*}
    \includegraphics[width=\textwidth]{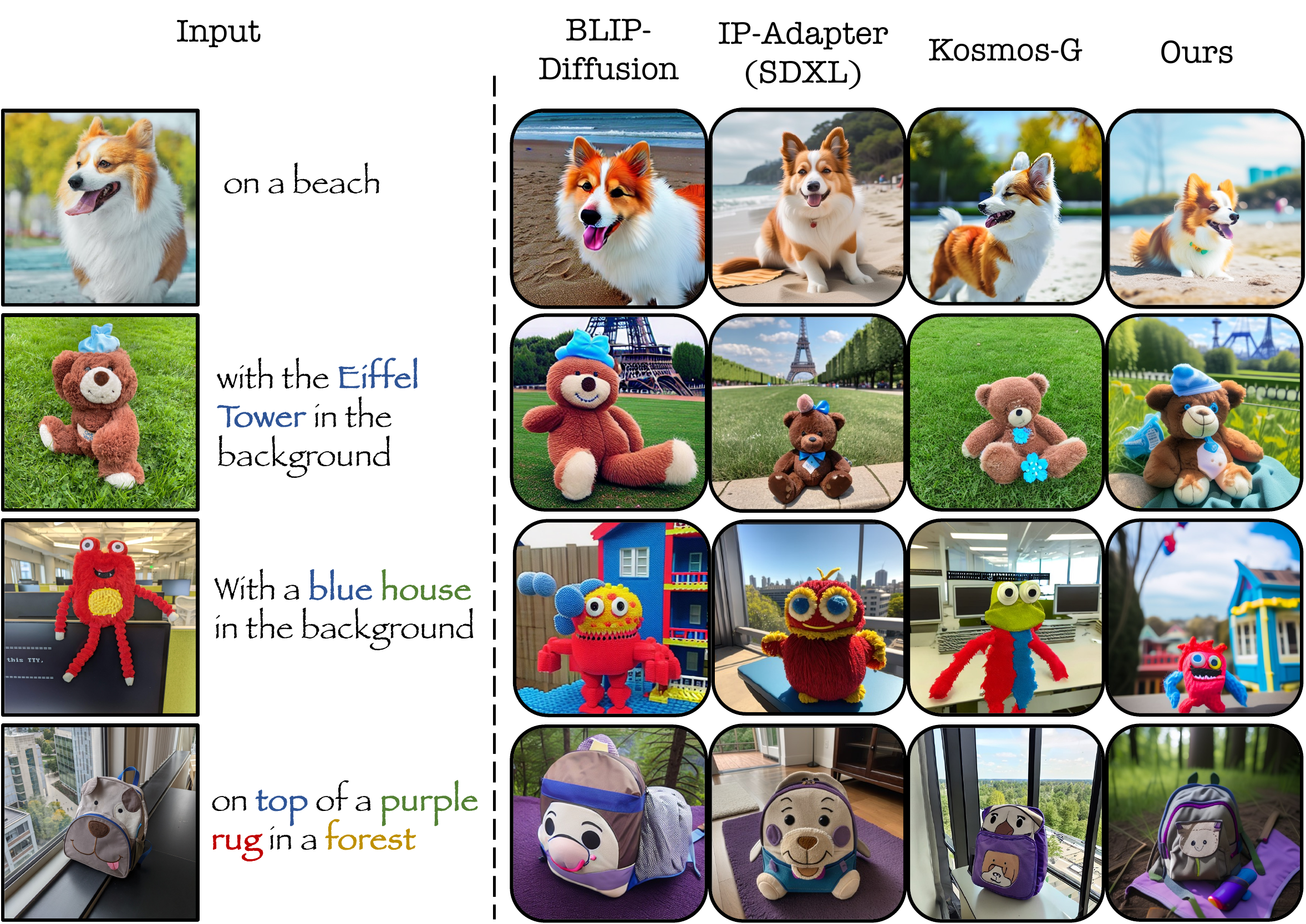}
    \caption{
     Qualitative examples of the increasing complexity of novel visual concepts as we move from top to bottom.
    }
    \label{fig:composition-cases}
\end{figure*}

In this section, we showcase a collection of detailed qualitative examples from the P-T2I generation process, highlighting the challenges of crafting complex compositions within \ours~and comparative models. 
As depicted in Figure~\ref{fig:composition-cases}, the complexity of the showcased examples progressively increases, illustrating a noticeable escalation in the intricacy of visual concepts from the top to the bottom of the figure. 
With the rising complexity, we note a universal decline in the ability of all methodologies, including \ours, to preserve subject fidelity accurately.
Interestingly, despite these challenges, \ours~demonstrates a better grasp of compositional integrity, unlike the baseline models which falter across all complexity levels.

\begin{figure*}
    \includegraphics[width=\textwidth]{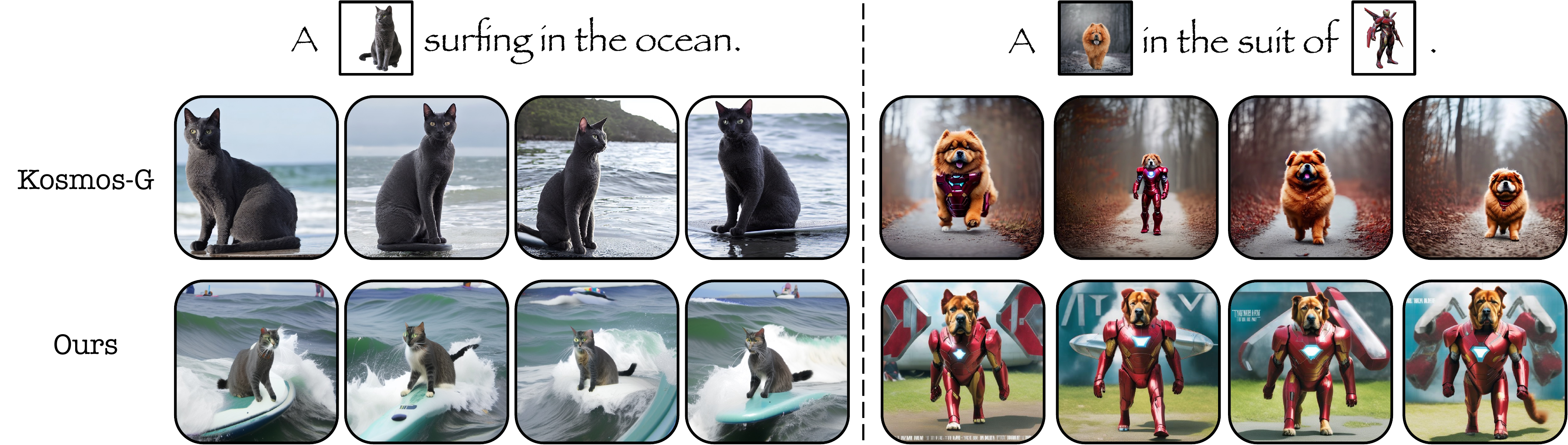}
    \caption{
     Qualitative examples of showcasing the consistency comparisons between Kosmos-G and \ours.
    }
    \label{fig:inconsistency-cases}
\end{figure*}

Moreover, we present instances demonstrating the variability in outcomes produced by P-T2I methods across different trials. 
As illustrated in Figure~\ref{fig:inconsistency-cases}, while there is a semblance of consistency in generating single and multiple concepts between models, Kosmos-G specifically shows variability in rendering multiple concepts—occasionally misplacing elements of the Ironman suit on a dog or failing to include it altogether. 
This phenomenon suggests that \ours~minimizes image diversity to enhance result consistency, a trait observed across the UnCLIP model family.

\begin{figure*}
    \includegraphics[width=\textwidth]{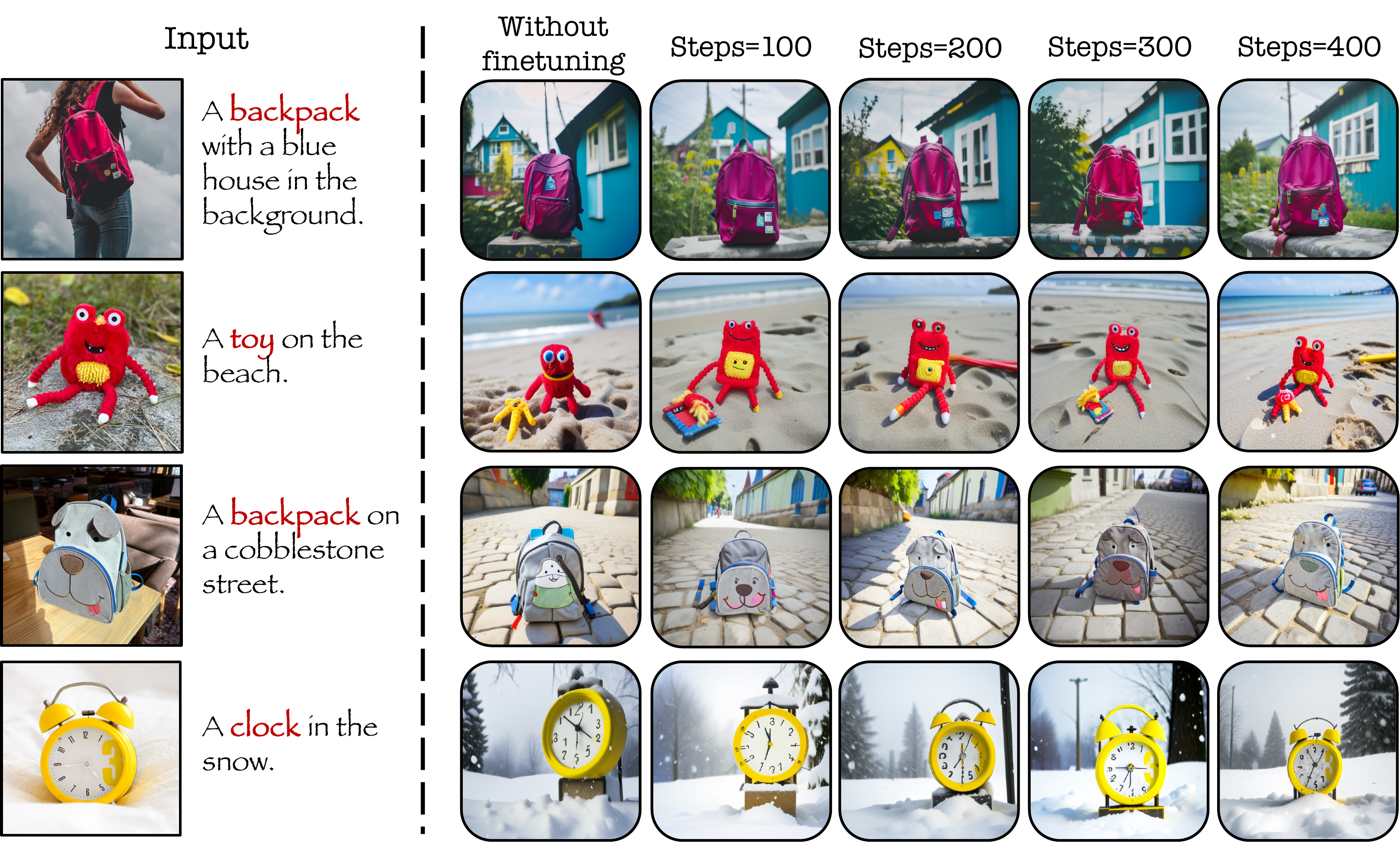}
    \caption{
     Qualitative examples of \ours~without finetuning and different stages of finetuning.
    }
    \label{fig:finetuning-examples}
\end{figure*}

Figure~\ref{fig:finetuning-examples} offers qualitative insights into the performance of \ours~without and with minimal fine-tuning. 
It is evident that in certain edge cases, where \ours~initially struggles to fully grasp novel visual concepts without finetuning, a modest application of few optimization iterations significantly enhances concept capture. 
Further optimization not only preserves text composition but also enriches minor, subject-specific details, underscoring the adaptability and finesse of \ours~in nuanced image generation.

\begin{figure*}
    \includegraphics[width=\textwidth]{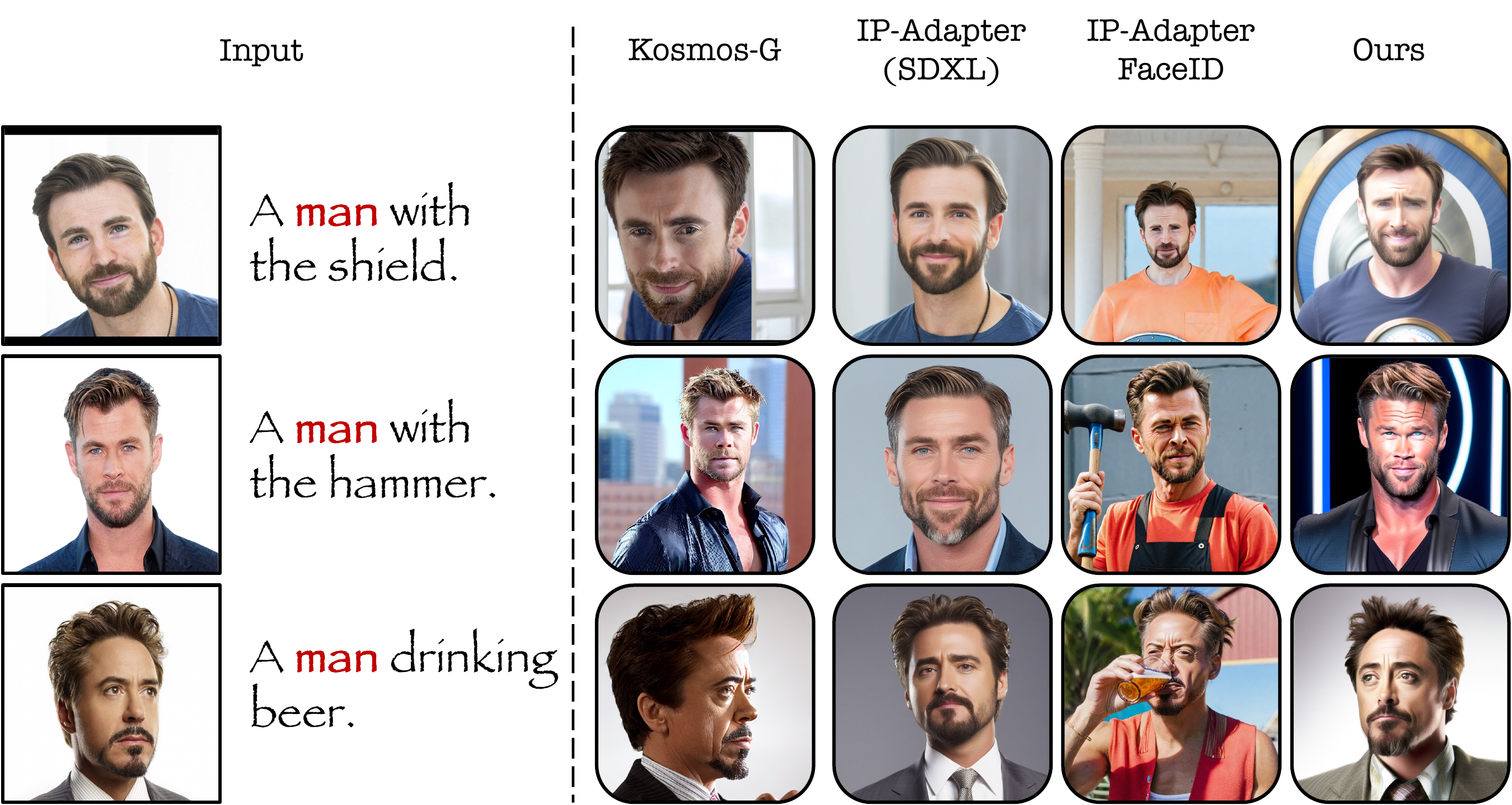}
    \caption{
     Qualitative examples of showcasing the failure cases on human faces on Kosmos-G, IP-Adapter (SDXL), IP-Adapter (FaceID), and \ours.
    }
    \label{fig:face-failure-cases}
\end{figure*}

\begin{figure*}
    \includegraphics[width=\textwidth]{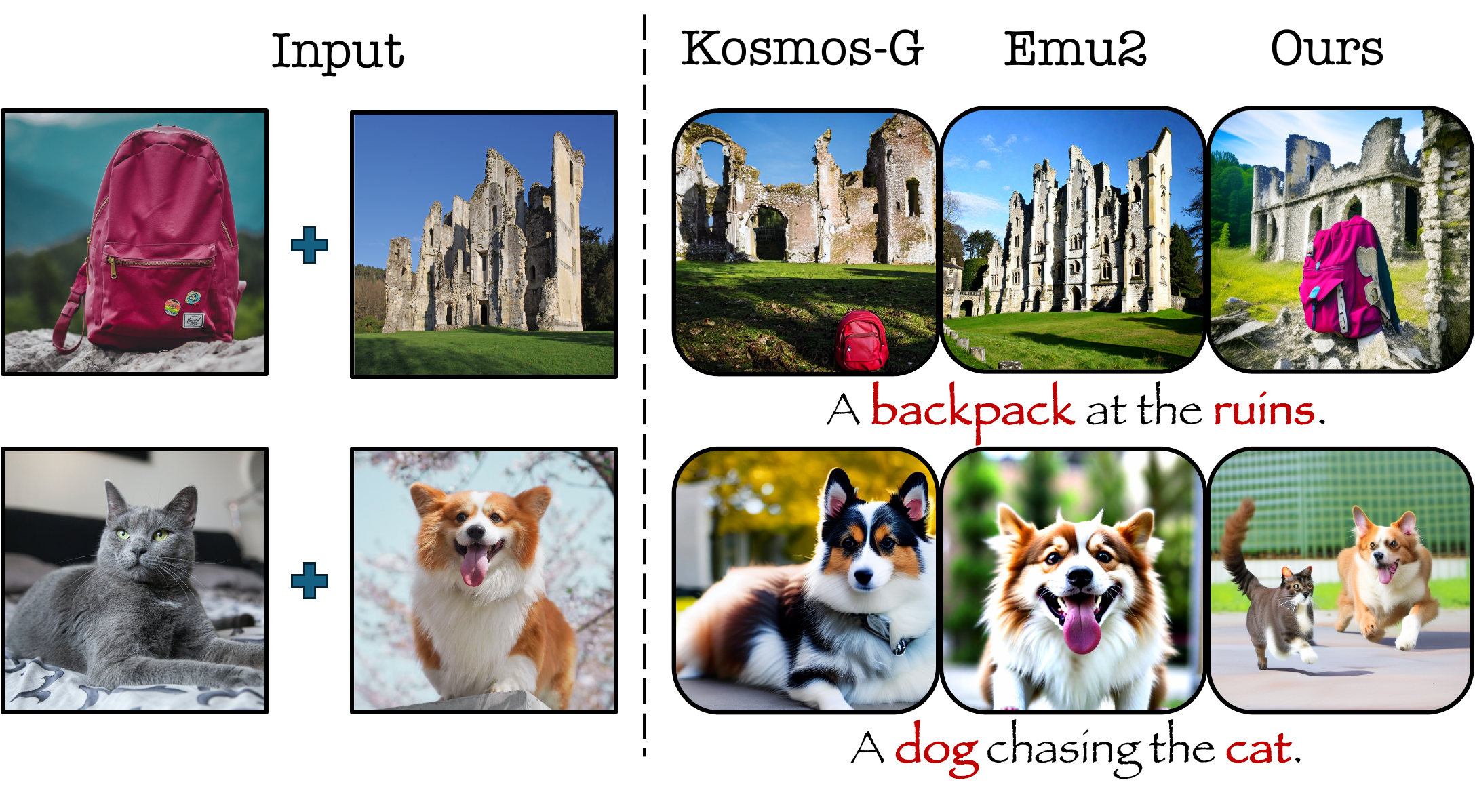}
    \caption{
     Qualitative examples of showcasing the failure cases on Multibench of Kosmos-G, Emu2, and \ours.
    }
    \label{fig:multibench-failure-cases}
\end{figure*}

Moreover, in our evaluations using the Multibench dataset, we noticed that both the baseline models (Kosmos-G and Emu2) and \ours~encounter difficulties in precisely maintaining all subject-specific details, as depicted in Figure~\ref{fig:multibench-failure-cases}. 
\textbf{This underscores that zero-shot multi-subject P-T2I generation remains a significant challenge in the field.}
Further, we explored how well each model preserves genuine human facial characteristics in various scenarios, particularly when combined with differing captions. 
The qualitative examples in Figure~\ref{fig:face-failure-cases} shed light on this aspect. 
Although each model strives to maintain the original facial features, none succeeds in replicating the specific personal facial details accurately. 
These instances typically fall short of precisely conveying the intended compositions, with the exception of one scenario in IP-Adapter FaceID, indicating a notable area for future improvements in model performance.

\section{Limitations}
Our work marks a pioneering venture into leveraging the latent space of pre-trained CLIP models for P-T2I generation. 
Nonetheless, it's crucial to recognize certain constraints. 
Primarily, despite its strengths, CLIP's inability to perfectly capture hierarchical representations occasionally leads to less-than-ideal results.
This issue, stemming from the CLIP contrastive loss, can cause deviations from the original subject features, particularly when generating P-T2I for complex subjects like human faces.
We believe that enhancing CLIP's representations could significantly boost our framework's efficacy in P-T2I mapping. 
The \ours~model, trained on 34 million parameters and 1.6 million images, presents a substantial foundation.
Yet, there's potential for further refinement, as increasing the quality of data and the number of parameters could yield even better outcomes.

\section{Broader Impact}
Subject-driven image generation or Personalized Text-to-Image (P-T2I) methods have the potential to be a transformative tool in numerous domains. For their positive influence, they enable users to effortlessly generate, modify, and synthesize original subjects into diverse environments, thereby enriching creative expression. On the other hand, the ease of altering and creating images raises concerns about the responsible use of this technology which requires significant ethical and legal considerations. Users must be acutely aware of being able to infringe intellectual property rights and create misleading or harmful content. We recommend developers provide a more secure way such as image attribution~\cite{kim2023wouaf} for end-users to ensure accountability for misuse of such models. As such, those employing subject-driven image-generation techniques should exercise careful judgment, ensuring that their work adheres to ethical standards and legal boundaries. 
It is imperative that the broader implications of this technology are considered, and that a commitment to responsible and conscientious use guides its application.

\end{document}